\documentclass{article}

\usepackage[final]{neurips_2025}

\usepackage[utf8]{inputenc} % allow utf-8 input
\usepackage[T1]{fontenc}    % use 8-bit T1 fonts
\usepackage{hyperref}       % hyperlinks
\usepackage{url}            % simple URL typesetting
\usepackage{booktabs}       % professional-quality tables
\usepackage{wrapfig}        % wrapping figures and tables
\usepackage{amsfonts}       % blackboard math symbols
\usepackage{nicefrac}       % compact symbols for 1/2, etc.
\usepackage{microtype}      % microtypography
\usepackage{xcolor}         % colors
\usepackage{pifont}
\usepackage{graphicx}
\usepackage{amsmath}
\usepackage{caption} 
\usepackage{subcaption}

\newcommand{\tick}{\ding{51}} % ✓
\newcommand{\cross}{\ding{55}} % ✗

\title{Orbis: Overcoming Challenges of Long-Horizon Prediction in Driving World Models}

\author{%
  Arian Mousakhan\thanks{Main Contributors.} \\
  \And
  Sudhanshu Mittal\footnotemark[1] \\
  \AND
  Silvio Galesso\footnotemark[1] \\
  \And
  Karim Farid\footnotemark[1] \\
  \And
  Thomas Brox\\
}

\begin{document}

\maketitle
\begin{center}
\vspace{-3em}
    University of Freiburg, Germany\\
  \texttt{\{mousakha,mittal,galessos,faridk,brox\}@cs.uni-freiburg.de}\\
\end{center}

\begin{abstract}
%World models are a promising path to build a simulator from data, enabling data-efficient exploration and decision making in the world that matters. 

%We notice that public driving world models built on video generation struggle with long-horizon rollouts. For better understanding, we develop a simple architecture where different choices can be systematically compared.

% In this paper, we show that public driving world models built on video generation have problems with long-horizon rollouts. 
% We investigate, why this problem arises and how it can be mitigated. 

% In this paper, we show that public driving world models have problems with long-horizon rollouts and capturing realistic driving behaviour. We build the models from the ground to study the best choice of design for driving world models.

Existing world models for autonomous driving struggle with long-horizon generation and generalization to challenging scenarios. In this work, we develop a model using simple design choices, and without additional supervision or sensors, such as maps, depth, or multiple cameras.
We show that our model yields state-of-the-art performance, despite having only 469M parameters and being trained on 280h of video data. It particularly stands out in difficult scenarios like turning maneuvers and urban traffic. We test whether discrete token models possibly have advantages over continuous models based on flow matching. To this end, we set up a hybrid tokenizer that is compatible with both approaches and allows for a side-by-side comparison. Our study concludes in favor of the continuous autoregressive model, which is less brittle on individual design choices and more powerful than the model built on discrete tokens. Project page with code, model checkpoints and visualization can be found here: \href{https://lmb-freiburg.github.io/orbis.github.io/}{https://lmb-freiburg.github.io/orbis.github.io}
% {Orbis-link}.
%We will open source our model upon publication.

% In particular, we test whether discrete token models possibly have advantages over continuous flow matching-based models. To this end, we set up a hybrid tokenizer that is compatible with both approaches and allows for a side-by-side comparison. Our study concludes in favor of the continuous autoregressive model, which is less brittle on individual design choices and more powerful than the model built on discrete tokens. It yields state-of-the-art performance with only 469M parameters and 280h of video data. It particularly stands out in difficult scenarios like turning maneuvers and dense urban traffic. Its training is straightforward without the need for additional supervision or sensors, such as maps, depth, or multiple cameras. We will open source our model upon publication.
%We hypothesize that the problems of previous continuous models are due to their generation of multiple frames in one junk rather than generating content in a frame-by-frame autoregressive mode.    

% hybrid tokenizer -- semantic, equivariance..
% frame-wise generative model
    % flexible length generation
% long-horizon generation
% diverse generation
% controllable
% training bugdet ? 
% fast inference ? 
% new evaluation benchmark
% new metric for ego-motion dynamics for unconditional driving world models. 
% open-source

\end{abstract}

\section{Introduction}

% about world models in general and driving world models is specific. 
Intelligent agents operate in complex environments by simulating plausible future states based on past observations. This capacity for imagination allows them to plan toward long-term goals~\cite{ha2018world, bar2024navigation}. Humans naturally acquire this ability through passive observation and minimal interactions, enabling them to adapt quickly to new and unseen scenarios.
Emphasizing the passive observation component of such world models has become particularly popular for the driving world, since large amounts of data exists for this domain. It is attractive to circumvent the manual setup of many perception components by learning the visual representation for decision making via the predictive loss of a world model.

% opening para about previous methods.. what main mode of failure. Stopping
Recent driving world models~\cite{gao2024vista, agarwal2025cosmos, hassan2024gem} built on video diffusion models~\cite{blattmann2023stable} have made major strides towards generating detailed content in high definition and at high frame-rates. 
However, Fig.~\ref{fig:baselines_limitations} highlights that these models only work well for few frames, especially in case of maneuvers that require generating new content, such as turning. 
% This indicates limitations in how these models capture state transitions -- the key feature of a world model. 
% We quantify this shortcoming via a metric that compares the trajectory of the real continuation with the trajectories of the generated content. Our evaluation reveals consistent problems with all public driving world models. 
Our evaluation, conducted on the generated videos as well as on the estimated ego-trajectories, reveals substantial limitations in how all public driving world models capture state transitions -- the key feature of a world model -- even though the quality of the generated videos is excellent at first.

% One lingering open question is whether world models should build on predicting discrete tokens (similar to LLMs)\cite{esser2021taming, chang2022maskgit, yu2023magvit} or if they should rather predict continuous outputs. The current trend goes towards continuous modeling in diffusion frameworks \cite{ho2020denoising, song2020denoising, song2020score, karras2022elucidating}. However, this could be the core of the observed problems. GAIA-1~\cite{hu2023gaia}, which was built on a quantized tokenizer, showed generated content that had no issues with turns and long roll-outs.

{While realistic video generation can be a valuable product, it is not the primary objective of world models. As demonstrated in a number of influential works (e.g. World Models~\cite{ha2018world}, Dreamer~\cite{hafner2019dream}, Cosmos~\cite{agarwal2025cosmos}, VJEPA-2~\cite{assran2025v}), their main purpose is representation learning, planning, and policy learning. 
The quality of state representations and the accuracy of next-state predictions are therefore paramount, with decoded videos serving primarily as an indicator, e.g. of whether the model can execute a turn stably. Long-horizon prediction serves as a more meaningful measure of how well the model captures state transitions, while generalization to complex scenarios reflects its ability to model diverse real-world dynamics. Existing models such as Vista and GEM, though conceived with planning as possible application, often fail in non-trivial yet routine situations. In this work we prioritize building a world model with robust state representation and dynamics that handle such cases effectively.}

Consequently, a relevant question is whether world models should rely on continuous-space latents or use discrete tokens (similar to LLMs)\cite{esser2021taming, chang2022maskgit, yu2023magvit} for representing world states.
The current trend for visual generation goes towards diffusion-based (continuous) models \cite{polyak2024movie, blattmann2023stable, girdhar2024factorizing}. On the other hand, driving world models based on discrete representations and LLM-like objectives seem to have the edge in terms of rollout duration~\cite{hu2023gaia, hu2024drivingworld}. Among these, the proprietary GAIA-1 model showed no issues with turns and long drives.
This observation prompts the question of whether the discrete paradigm is really superior to the other for long-term generation, and whether the continuous space is the reason for the observed shortcomings in the current state of the art.

% \textit{Is world modeling in the continuous space the reason for the observed shortcomings and can we fix this?}
\begin{wrapfigure}{r}{0.5\textwidth}
  \centering
  \includegraphics[width=\linewidth]{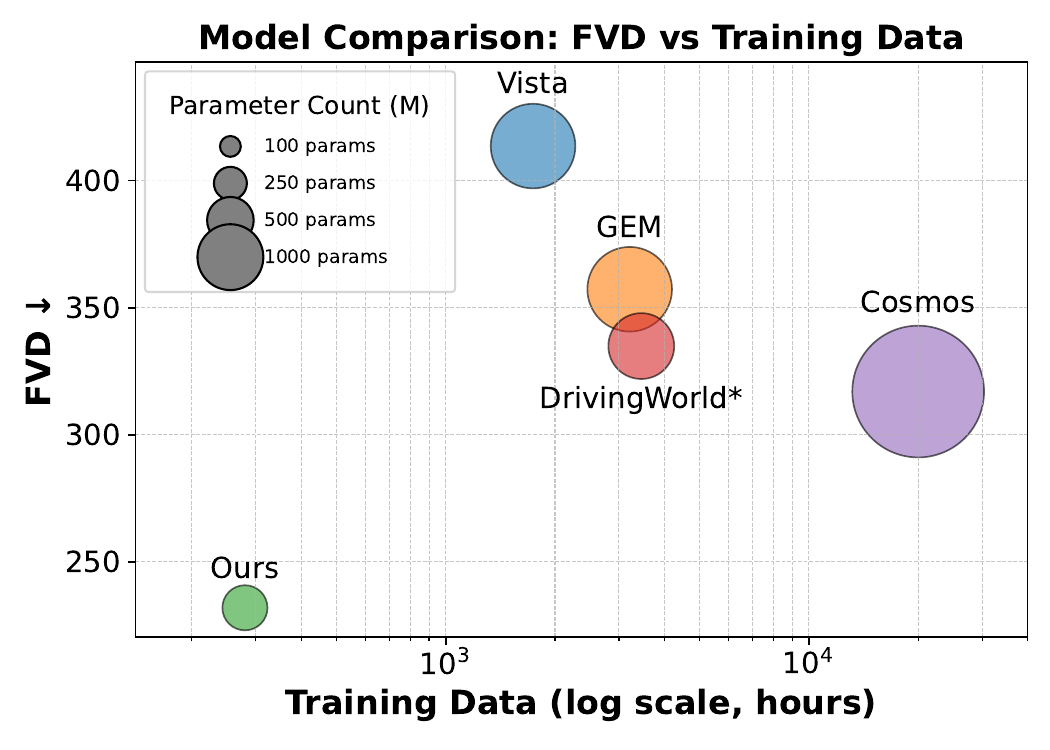}
    \caption{Comparison of model scale, training data volume, and FVD performance of various approaches on the NuPlan-turns dataset. *DrivingWorld is trained on the test dataset nuPlan.}
  \label{fig:bubblechart}
\end{wrapfigure}
%is trained on the test dataset nuPlan and uses ego-motion control as an extra input.

% Hypothesis: discrete vs. continuous
% To investigate the hypothesis that continuous video diffusion models are inferior to discrete models in terms of modeling state transitions. 
To address these questions, we introduce a hybrid discrete-continuous tokenizer that is compatible with both types of modeling approaches to be able to compare the two strategies on the same ground. For the quantized-token model we developed a frame-wise autoregressive model based on MaskGIT~\cite{chang2022maskgit}, whereas for the continuous-token model we developed an autoregressive model based on flow matching~\cite{lipman2022flow, liu2022flow}. Both models were trained from scratch. We also put effort into optimizing the details of the tokenizer. Indeed we find that many of these details are important for the performance of the model acting on quantized tokens. Surprisingly, these details are of little relevance for the continuous modeling approach. Both our models can handle long roll-outs, but the continuous approach yields significantly better results and sets the state of the art by a large margin; see Fig.~\ref{fig:bubblechart}. 

% now more in detail about them 
% strong priors
%Driving world models typically follow a two-stage modeling approach, similar to video generative models. The tokenizer model encodes visual inputs into compact latent representations and a latent world model learns the transition dynamics between these latent states.
%Multiple driving world models leverage general-purpose pre-traiend tokenizers~\cite{sd} to encode the input observations and rely on video generative models to initialize the latent world model. However, we observe that the representations produced by such general-purpose tokenizer are suboptimal for learning a driving world model. Similarly, strong priors learned by generic video generation models often produce infeasible motion, like horizontal drifting, likely due to a distribution mismatch. \todo{show some weird trajectories in the teaser.} Despite using a strong prior, latent world models are often trained with thousands of hours of driving videos. In this work, we present a domain-targeted tokenizer and world model that achieves superior performance while requiring significantly less model capacity and training data.

% extra conditioning
% less focus on representation learning
%While many prior works have focused on advancing latent world models through different objectives, conditioning and regularization techniques, the importance of the tokenizer's latent representation has been overlooked. 
%In this work, we improve the representation produced by the tokenizer model through factorized tokenization and enforcing scale equivariance. 

Unlike many prior approaches, our world model is trained using only raw video data without using any extra low-level regularization objectives, such as structural consistency or pseudo-depth supervision. All implicit perception is learned directly from the presented videos. 
% This makes the approach perfectly scalable.
This makes the approach more scalable and establishes a strong foundation for the development of more controllable models.

% \todo{Emphasize the simplicity of our model. Prior work use strong priors and implicit low-level regularizers like structural consistency or pseudo-depth supervision in the latent world model training.}
We also demonstrate that our model can be modified easily to allow ego-motion control via adaptive layer normalization~\cite{perez2018film}. To this end, we evaluate the trajectories produced by our world model also in ego-motion-control-conditioned settings, where we propose a set of metrics to evaluate realism and coverage of the requested trajectories. 

% \todo{add about controllability}

To summarize, (1) we highlight shortcomings of contemporary driving world models and propose additional benchmarking metrics to make these shortcomings more explicit. 
(2) We propose a hybrid discrete-continuous tokenizer that is compatible with both discrete and continuous prediction losses and allows us to compare both modeling approaches side-by-side. 
(3) On its basis, we compare continuous and discrete prediction losses on a fair common ground and find a clear advantage in favor of continuous modeling. 
(4) As demonstrated in Figure~\ref{fig:bubblechart}, the resulting model is much more economical in terms of training data and model size than existing world models. Using only 280 hours of front camera video data, our 469M parameter flow-matching model Orbis already produces state-of-the-art performance on long-horizon rollouts with realistic and diverse trajectories. It excels particularly in challenging driving scenarios. 
\section{Related Work}

% world models in abstract space do not have environment information to react. Relies on heavy annotations

%\paragraph{World models} % ways of world modeling
% previous works -- non-video modelling based

\textbf{World Models.} The ability of world models to do real-world simulation can be useful for policy learning~\cite{piergiovanni2019learning, he2024large}, sample efficient RL~\cite{hafner2019dream, hafner2020mastering, wu2023pre, li2024think2drive}, and representation learning~\cite{zhao2024drivedreamer4d}. Previous world models have been limited to gaming~\cite{hafner2020mastering, hafner2019dream, ha2018recurrent} and other simulated environments~\cite{dosovitskiy2017carla}.
Recent breakthroughs in video generative modeling~\cite{yang2024cogvideox, blattmann2023stable} have led to future video prediction models - an essential building block for world models. 

% sora ?? 

% Given the complexity of driving scenarios, many models based on video generative models use available annotations to learn egocentric dynamics and control the generation process. 
Multiple driving world models~\cite{wang2024drivedreamer, Wen_2024_CVPR, li2024bevformer} use BEV (Bird’s-Eye-View) annotations like depth maps, 3D bounding boxes, road maps to generate new scenarios. 
DriveDreamer~\cite{wang2024drivedreamer} incorporated multi-modal input, such as traffic conditions, text prompts, and driving actions, for future frames and action generation. 
Many other works~\cite{Wen_2024_CVPR, zhao2024drive, wang2024driving} extended this idea to multi-view video generation. 
Some recent works~\cite{zhao2024drive, jia2023adriver} also use LLMs and VLMs~\cite{liu2023visual} for spatial reasoning. 
Although these models show high quality generation, they rely on heavy external knowledge. Such heavy reliance limits the model's ability to generalize to new environments. In this work, we train a generalizable world model using unannotated front-camera videos and only fine-tune for ego-motion control. % later to add steering ability.

Recent driving world models~\cite{kim2021drivegan, yang2024genad, gao2024vista, agarwal2025cosmos, hassan2024gem, hu2023gaia, russell2025gaia} trained predominantly on raw driving video data have shown the ability to simulate realistic future scenes in unseen environments. 
DriveGAN~\cite{kim2021drivegan}, among the first works to train on real-world driving data, showed realistic future generation with ego-motion and environment controllability.
GAIA-1~\cite{hu2023gaia} further enhanced the quality of future prediction and added controllability through text, in addition to action input.  %Vista~\cite{gao2024vista} shows 
Diffusion-based world models~\cite{gao2024vista, hassan2024gem} fine-tuned general-purpose pre-trained video generation models like SVD~\cite{blattmann2023stable} to produce future video predictions at high resolution and high frame rate. Driving world models - Vista~\cite{gao2024vista} and GEM~\cite{hassan2024gem} demonstrate high-quality rollouts up to 15 seconds. DrivingWorld~\cite{hu2024drivingworld} further enables longer and more coherent rollouts.

% at 576x1024 resolution at 10Hz. GEM further added controls to alter the scene objects along with ego-motion control.

Generative models based on vector-quantized tokens like autoregressive~\cite{yan2021videogpt, 10.5555/3692070.3693075} and masked generative models~\cite{Yu_2023_CVPR, liu2024mardini, gupta2022maskvit}, have also demonstrated strong performance in video generation due to their strong capability in modeling dynamics and representation learning.
For world modeling, Genie~\cite{bruce2024genie} and GAIA-1~\cite{hu2023gaia} have demonstrated generalized world modeling capabilities with interactive control and long-horizon rollouts respectively. We also observe that quantized driving world model can perform long-horizon generation. %However, they fail to capture small movements reducing model's expressivity.

\textbf{Latent Representation Learning.}
VAEs~\cite{pinheiro2021variational} and VQVAEs~\cite{van2017neural,esser2021taming} are foundational autoencoding techniques for learning latent representations used in training latent world models. VAEs produce the continuous latents, commonly used in diffusion~\cite{zhou2022magicvideo, blattmann2023stable} and flow matching~\cite{russell2025gaia, polyak2024movie, jin2024pyramidal} generative models. VQ-VAE produces discrete (quantized) latent codes for LLM-style autoregressive~\cite{esser2021taming, kondratyuk2023videopoet} and masked generative modeling~\cite{Yu_2023_CVPR, bruce2024genie}. Following VQGAN work~\cite{esser2021taming}, ideas such as product quantization~\cite{li2025imagefolder, qu2024tokenflow, bai2024factorized}, residual quantization~\cite{lee2022autoregressive}, multi-scale residuals~\cite{tian2024visual, ma2024star}, spectral decomposition~\cite{esteves2024spectral, lin2024open, agarwal2025cosmos} have been introduced for image and video generation. Some works also propose hybrid tokenizers~\cite{tang2025hart} that unify the tokenizer model for both discrete and continuous generative models. %This design is for their second stage to initially predict discrete tokens and then map it to continuous space. 

\begin{figure}[!t]
  \centering
  \includegraphics[width=1.0\linewidth]{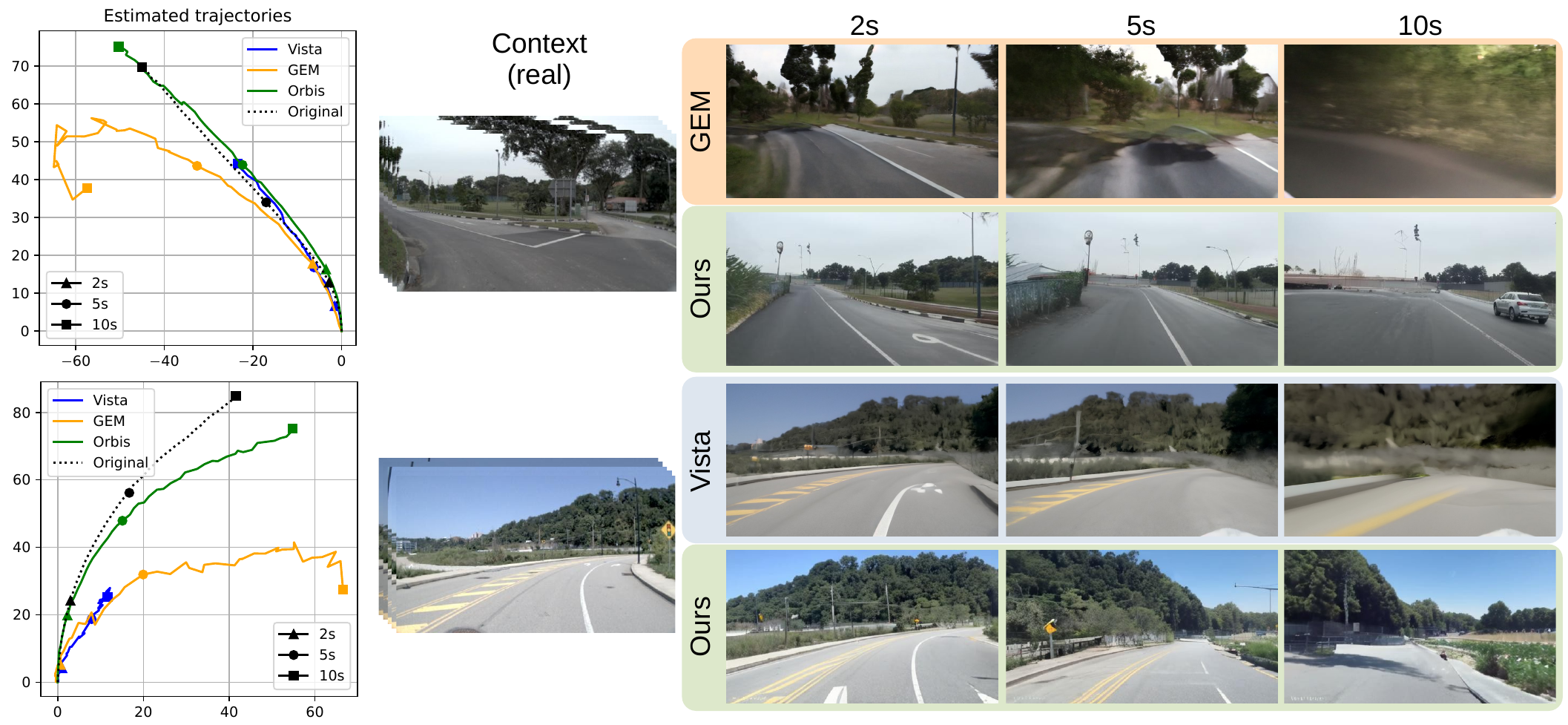}
  \caption{Limitations of state-of-the-art video generation models on turning events. \textbf{Left}: The trajectories estimated from the generated videos show that previous approaches either stop prematurely or drift into an unnatural path. \textbf{Right}: The quality of the corresponding generated frames degrades over time, as the models struggle to generate the scenery. In contrast, our method tracks the original trajectory curvature and speed more closely, and can generate novel content beyond the unseen horizon. Videos for a larger set of randomly sampled context frames are linked in the Appendix.}
   \label{fig:baselines_limitations}
\end{figure}

\section{Challenges Faced by Contemporary Driving World Models} 
% another option -- When and Why Baselines Struggle

% \paragraph{Strong priors hinder realistic driving motion.}
% % horizontal jitters
% % Ignoring context. For eg ignore turning context and continue straight.
% Many previous works rely on pre-trained tokenizers and video generation models trained on general purpose datasets. 
% While these driving world models are fine-tuned on thousands of hours of driving data, we suspect that the strong mismatched priors from pre-training effect the smooth long motion of the vehicle. It leads to unrealistic ego-vehicle behaviors, such as horizontal sliding or jitter.
% Such artifacts can be seen in the trajectories produced by previous models, as shown in Fig.~\ref{fig:baselines_limitations}.
% We also observe that during turning events, these models tend to ignore context frames and continue forward motion, likely due to a mismatched prior.

Prior works perform well in straight-road driving scenarios but show significantly higher failure rates when faced with difficult maneuvers. For example, in turning events, as rollouts extend over longer horizons, the content generated by these works tends to run out of distribution, producing blurred frames. 
% The degradation of visual semantics and details causes the ego-vehicle to stop prematurely, as the model fail to pick up reliable condition signals from its own generated frames.
The degradation of visual semantics and details causes the ego-vehicle to stop prematurely, as the model cannot recover from poorly generated new context.
This can be seen in Figure~\ref{fig:baselines_limitations}, where Vista~\cite{gao2024vista} comes to a halt before 10 seconds. Most of these limitations can be reflected using video quality metrics like FVD~\cite{unterthiner2019fvd} computed on several time windows, as shown in Section~\ref{sec:res}.

% \begin{figure}[!ht]
%   \centering
%   \includegraphics[width=0.9\linewidth]{figures/teaser_img.pdf}
%   \caption{Limitations of state-of-the-art video generation models on turning events. \textbf{Left}: The trajectories estimated from the generated videos show that previous approaches either stop prematurely or drift into an unnatural path. \textbf{Right}: The quality of the corresponding generated frames degrades over time, as the models struggle to generate the scenery. In contrast, our method tracks the original trajectory curvature and speed more closely, and can generate novel content beyond the unseen horizon.}
%    \label{fig:baselines_limitations}
% \end{figure}

We found that these models also show unrealistic ego-vehicle behaviors, such as lateral sliding or jitter artifacts. This could be due to the strong priors inherited from general-purposed pre-trained video generation models.
% We speculate that these are caused by the strong priors inherited from general-purposed pre-trained video generation models.
Such artifacts can be seen in the trajectories produced by these models, as shown in Fig.~\ref{fig:baselines_limitations}.

These are not well captured by standard metrics like FVD or JEDi, whose reliance on pretrained human-action or general purpose encoders limits sensitivity to driving dynamics and ego-motion. This calls for more targeted evaluations. To bridge this gap, we propose a distribution-level trajectory-based evaluation, detailed in Section~\ref{sec:traj}, that directly quantifies realism and coverage of generated driving behavior compared to a curated dataset of turning events.
We evaluate and compare the generated trajectories for Vista, GEM, and our approach, and find the results to confirm our qualitative observations and show the shortcomings of the existing methods.

\section{Compatible Discrete and Continuous Prediction Models}% via a Hybrid Tokenizer}

The above shortcomings all appear in conjunction with approaches based on video diffusion. This modeling approach could be a potential cause of these methods' failure. 
To enable a fair comparison between discrete and continuous latent world models, we design a hybrid image tokenizer that supports both objectives and allows us to evaluate directly which objective better handles the challenges of long-horizon prediction in a simple and controlled setting.
Our study is conducted using two efficient formulations: flow matching for continuous models and masked generative modeling for discrete models. 
% Additionally, we analyze how different tokenizer design choices impact the performance of the dynamics world model.

%Similar to Vista, GEM, GAIA-1, we use frame-wise tokenizer model. 

\subsection{Hybrid Image Tokenizer}
\label{section:tokenizer}
\paragraph{Preliminary.} 
% A visual tokenizer aims to compress high-dimensional pixel data into a low-dimensional latent representation. VQ-VAE~\cite{van2017neural} achieves this by encoding the image through an encoder–decoder architecture and applying a vector-quantization (VQ) bottleneck. 
Given an image $\mathbf{I} \in \mathbb{R}^{H \times W \times 3}$, the encoder $\mathcal{E}$ produces a latent $\mathbf{x}=\mathcal{E}(\mathbf{I}) \in \mathbb{R}^{H' \times W' \times d}$, where d is latent channel dimension. 
The latent $\mathbf{x}$ is then quantized to the closest codebook entry, resulting in $\mathbf{q}=\mathcal{Q}(\mathbf{x}) \in \mathbb{R}^{H' \times W' \times d}$, using a codebook $\mathcal{C} \in \mathbb{R}^{K \times d}$ with $k$ entries. The tokenizer is trained using the VQGAN~\cite{esser2021taming} objective.

\begin{figure}
  \centering
  \includegraphics[width=1\linewidth]{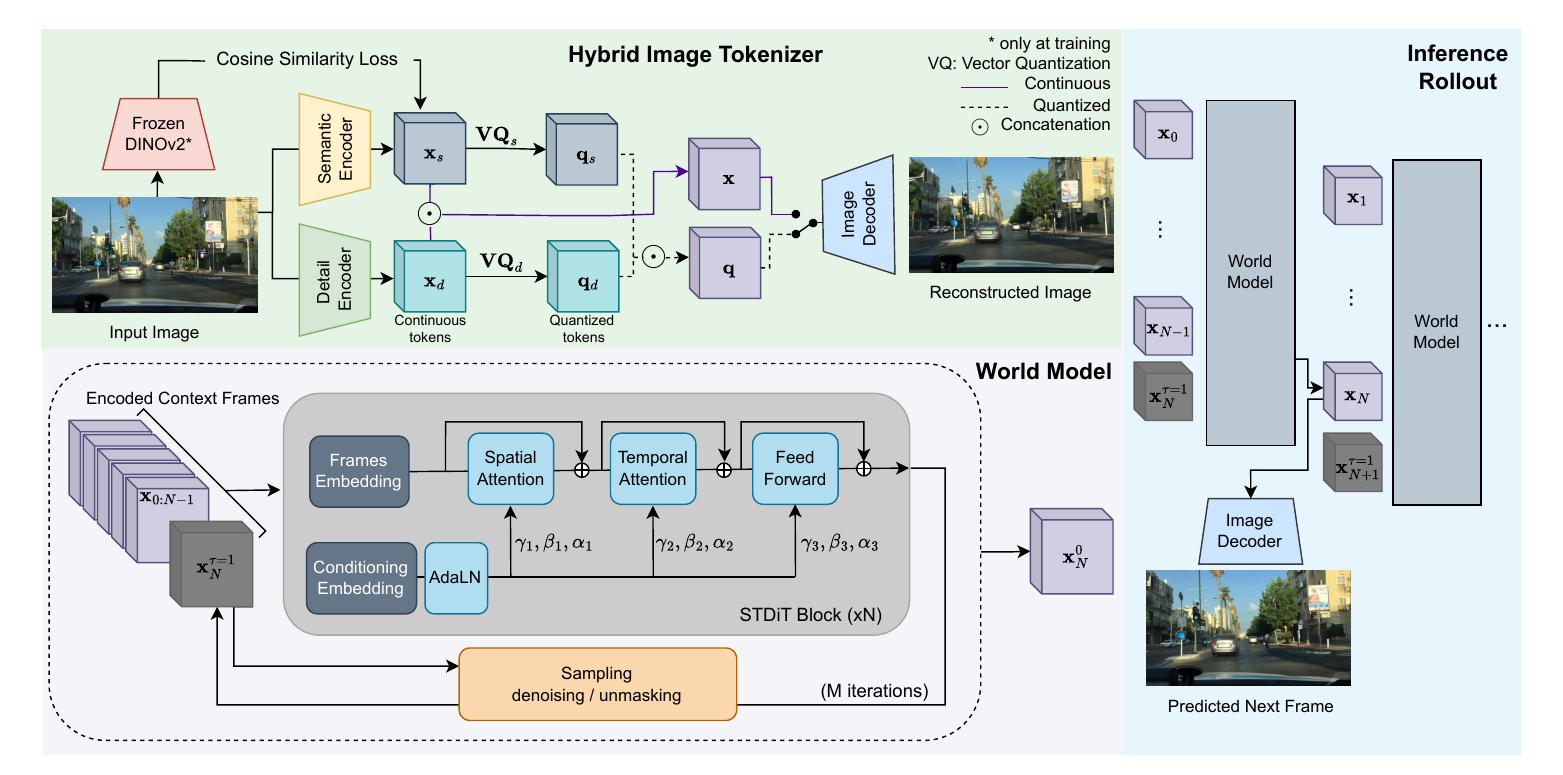}
  \caption{\textbf{Image Tokenizer:} The tokenizer provides two semantic and detail representation. These two representations are concatenated and fed into the image decoder and later to the world model. During training the decoder receive continuous or discrete tokens randomly in the fine-tuning phase. \textbf{World Model:} To generate the next frame, the model receives either sampled Gaussian noise or fully masked tokens as the target frame, along with encoded context frames. The model progressively denoise or unmask the target frame. This iterative sampling process is repeated to generate target frame. \textbf{Inference Rollout:} During inference, the world model autoregressively generates next frame. This process repeats for the desired number of frames in the rollout sequence.
  }
   \label{fig:framework}
\end{figure}

\textbf{Our design.} 
Building upon recent works~\cite{li2025imagefolder, yu2021vector, yu2023language}, we design a custom hybrid tokenizer, suitable for both discrete and continuous predictive video modeling. 
VQ-VAEs~\cite{van2017neural} typically optimize latent representation learning for pixel-level reconstruction. 
%Prior works show that their learned latents often fail to support generative modeling effectively. 
Prior works show that these representations typically lack desirable properties such as semantic structure~\cite{li2025imagefolder}.% and transformation equivariance~\cite{what?}. 
To address these limitations, we adopt a factorized token design~\cite{bai2024factorized}, using separate encoders \(\mathcal{E}_s\) and \(\mathcal{E}_d\) to produce \textit{semantic} tokens \(\mathbf{x}_s\) and \textit{detail} tokens \(\mathbf{x}_d\), respectively. The former are obtained via additional distillation from DINOv2~\cite{oquab2023dinov2} as shown in the Figure~\ref{fig:framework}.
Each output is quantized independently using separate codebooks \(\mathcal{C}_s\) and \(\mathcal{C}_d\), yielding \(\mathbf{q}_s = \mathcal{Q}_s(\mathbf{x}_s)\) and \(\mathbf{q}_d = \mathcal{Q}_d(\mathbf{x}_d)\). 
We convert the tokenizer into a \textit{hybrid model} by fine-tuning it with a 50\% probability of bypassing the VQ bottleneck during training. This simple modification allows a single tokenizer to support both discrete and continuous latent representations. The final continuous representation is $\mathbf{x}=(\mathbf{x}_s; \mathbf{x}_d)$ and the corresponding quantized representation is $\mathbf{q}=(\mathbf{q}_s; \mathbf{q}_d)$.
The decoder reconstructs the image from the final representation.

%The final representation 
%Given continuous or discrete representation the decoder is trained to reconstruct the image.
% The decoder receives concatenation of both detail and semantic embedding to reconstruct the image.

%After training the tokenizer, we noticed a large portion of the codebook was not utilized.
% To address codebook under-utilization issues of VQGAN~\cite{esser2021taming} we incorporate L2-normalized codes~\cite{yu2021vector}, low-latent dimension~\cite{yu2021vector} and entropy penalty~\cite{yu2023language}.
% To further improve the tokenizer, we fine-tune the model with implicit regularization as proposed in EQ-VAE~\cite{kouzelis2025eq}. The tokenizer is trained without the adversarial loss, similar to Cosmos~\cite{agarwal2025cosmos}, and only the decoder is fine-tuned later with the adversarial loss for improving the image quality. 
% , encouraging the decoder output \(\mathcal{D}(\text{Down}_s \cdot \mathcal{E}(\mathbf{I}))\) to approximate the downscaled input \(\text{Down}_s \cdot \mathbf{I}\), where \(\text{Down}_S\) denotes a downscaling factor (e.g., \(2\times\)–\(4\times\)). 

\subsection{Latent Space World Model}
% this goes in the model intro
%The goal of the dynamics model is to learn spatio-temporal dynamics representations for predicting the next frame given the past observation. 

We formulate our world model as a next-frame autoregressive model for both discrete and continuous objectives, as demonstrated in Figure~\ref{fig:framework}. 
%The context frames undergoes encoding process, as explained in section ~\ref{section:tokenizer}, to provide the latent representation for the world model.
%During training, the context of length $N-1$ frames is encoded frame-wise using the image tokenizer. The model also receives a partially noised version $\mathbf{x}^{\tau}_N$ of the $N^{th}$ frame.
%Given the input $\mathbf{x}_{0:{N-1}}  \oplus \mathbf{x}^{\tau}_N $, the objective of the dynamics model is to predict the denoised next frame $\mathbf{x}_N$, either  implicitly or directly, based on the formulation.
The model receives the context frames $\mathbf{x}_{0:N-1}$ and a target frame $\mathbf{x}^{\tau=1}_{N}$, initialized by noise or a complete mask. The model predicts the next frame $\mathbf{x}^{\tau=0}_{N}$ iteratively over multiple $M$ denoising or unmasking steps.
During inference rollouts, the model updates its context by appending the most recently generated frame $\mathbf{x}^{\tau=0}_{N}$, discarding the earliest context frame $\mathbf{x}_{0}$ of the previous inference step. This sliding-window process is repeated for each next-frame generation to get long-horizon predictions in the latent space (Figure~\ref{fig:framework}).
For visualization, each generated latent is decoded into an image using the tokenizer decoder.% model.

%As the efficiency in world model plays a crucial rule, 
In this work, we consider the flow matching ~\cite{lipman2022flow, liu2022flow, esser2024scaling} objective for the continuous world model and the masked generative modeling objective~\cite{chang2022maskgit, Yu_2023_CVPR} for the discrete world model.

%Given this representation, the objective of the dynamics model is to predict the next frame $\mathbf{x}_N$ conditioned on context frames $\{\mathbf{x}_0, \mathbf{x}_1, \ldots, \mathbf{x}_{N-1}\}$, where N is the number of frames. 
%Once the model learn to predict next frame representation, we leverage the model to rollout longer frames. 
%In order to rollout, the model receives the context frames $\mathbf{x}_{0:N-1}$ and initialized target frame, by noise or mask, $\mathbf{x}_{N}$. 
%Hence the model iteratively over M steps sample less corrupted target frame. 
%Once the target Frame is generated after M steps, we drop the first context frame and append the generated target frame to the context frame and initialize the next target frame. 
%For visualization each generated latent target frame is passed to the image decoder.

\subsubsection{Flow Matching }
%To formulate dynamics world model in a stable and efficient manner in continuous latent space, 
We follow the flow matching (FM) objective introduced by Lipman ~\cite{lipman2022flow}: we define a forward trajectory from the data distribution to a standard normal distribution via linear interpolation:
\begin{equation}
    \mathbf{x}^\tau = (1-\tau)\mathbf{x} + \tau\epsilon,\quad \tau \in [0,1],\quad \epsilon \sim \mathcal{N}(0, I)
\end{equation}
To use the flow matching objective for next frame prediction, the corrupted target frame $\mathbf{x}_{N}^{\tau}$ is conditioned on previous frames $\mathbf{x}_{0:N-1}$. The model predicts $\mathbf{v}(\mathbf{x}_N^{\tau}; \mathbf{x}_{0:N-1})$, the velocity that would take $\mathbf{x}_N^\tau$ towards the Gaussian prior. We train the model with the following objective as:
\begin{equation}
    \mathcal{L}_{} = \mathbb{E}_{\tau\sim[0,1],\,\epsilon\sim\mathcal{N}(0,I)}
 \big[ ||\mathbf{v}(\mathbf{x}_N^{\tau}; \mathbf{x}_{0:N-1})-(\epsilon-\mathbf{x}_N)||^2 \big],
\label{eq:flow_loss}
\end{equation}
At inference time, we sample a noise vector as the new target frame and iteratively transform it towards the data manifold. At each iteration the model calculates the velocity conditioned on context frames and update the target frame as:
\begin{equation}
    \mathbf{x}_N^{\tau-\delta_\tau} = \mathbf{x}_N^{\tau} - \delta_\tau \cdot \mathbf{v}(\mathbf{x}_N^{\tau}; \mathbf{x}_{0:N-1})
\end{equation}

where $\delta_\tau$ is the step size used to update the target frame at time step $\tau$. After integrating from $\tau = 1$ to $\tau = 0$, the resulting latent $\mathbf{x}_N^0$ is the generated next frame in latent space.

\subsubsection{Masked Generative Model }
In the discrete setting, we extend masked generative modeling (MGM), following the MaskGIT objective~\cite{chang2022maskgit}, from image generation to next-frame prediction. 
%We adapt masked image modeling into a next-frame prediction task. 
%Specifically, after encoding each frame into the continuous latent space \(\mathbf{x}\) (as used in the flow matching objective), we quantize the latent representation into a sequence of discrete tokens \(\mathbf{q}\).
The encoded latents of image frames are represented using discrete tokens $\mathbf{q}$. 

During training, we apply a binary mask \(\mathcal{M} \in \{0, 1\}^{H' \times W'}\) to the target frame \(\mathbf{q}_N\), resulting in the masked frame 
\(\mathbf{q}_N^{\mathcal{M}} = \mathbf{q}_N \circ \mathcal{M} + \texttt{[MASK]} \circ (1 - \mathcal{M})\), where \(\texttt{[MASK]}\) is a learned special token. 
The masking ratio for the whole frame is sampled uniformly from 0\% to 100\% using a predefined scheduler.
The MGM model takes as input the concatenated sequence of context frames and the noised target frame $(\mathbf{q}_{0:N-1}; \mathbf{q}^\mathcal{M}_N)$, and is trained to predict all the token IDs of the target frame. The training objective is a standard cross-entropy loss, defined as follows:

\begin{equation} 
    \mathcal{L}_{\text{CE}} = 
\mathbb{E}_{\mathcal{M}} 
\left[
- \sum_{i} \log p_\theta\left( \mathbf{q}_N^{(i)} \mid \mathbf{q}_{0:N-1}, \mathbf{q}_N^{\mathcal{M}} \right)
\right],
\end{equation}

where \(i\) indexes over all token positions in the target frame \(\mathbf{q}_N\).
%The model takes as input discrete token indices from the tokenizer, 
The model receives discrete token indices from the tokenizer, which discards any pairwise similarity structure among latent tokens. Following~\cite{savov2025exploration}, to reintroduce this structure, we utilize the similarities between quantized code vectors in the VQ codebooks as an extra regularizer to improve the training objective. 
At inference, given a fully masked target frame and the context frames $(\mathbf{q}_{0:N-1})$, the model iteratively predicts and replaces masked tokens. We follow, the confidence-based sampling~\cite{chang2022maskgit} heuristics for unmasking the target frame. %, where multiple tokens are unmasked in each step based on the confidence. % (e.g., based on maximum softmax probability). 

\subsubsection{Conditioning with Ego-motion}
To verify that our model is capable of action control, we implement the option for additional condition signals via adaptive layer normalization~\cite{perez2018film}. We embed steering angle and speed with a two-layer MLP, and add them to the other condition signals.
% \todo{more needed?}
% Notably, we don't train our models with action conditioning by default: the feature is present only when explicitly stated.

% \paragraph{Refinement stage} -- depends on the results plot. 

%At each iteration, it computes token probabilities for all masked positions and selects the most confident tokens (e.g., based on maximum softmax probability). 
%These tokens are then used to update the target frame. This process continues until all target frame tokens are unmasked.

%Stage-3: FM based-video decoder
%\subsection{Refinement Stage}

\section{Experiments}

\subsection{Experiment Details}

\paragraph{Datasets.} 
% OpenDV - 5670088 @10hz frames - 157.5 hours \\
% BDD100k - 4019033 @ 10hz  frames - 111.6 hours

\begin{wraptable}{r}{0.4\textwidth}
\small
  \centering
  \caption{Overview of the training datasets used for the world model.}
  \vspace{2 mm}
  \label{tab:dataset}
  \begin{tabular}{lccc}
      \toprule
            & Total & Used & Frames\\
      Name    &  (h) & (h) & (M) \\
      \midrule
      OpenDV  & 1747        & 158            & 5.67    \\
      BDD100K     & 1000        & 112            & 4.02 \\
      \midrule
      Total        & 2747  & 280 & 9.69  \\
      \bottomrule
    \end{tabular}
  \vspace*{0.5cm}
\end{wraptable}

To train our world model, we use subsets of videos from the BDD100K~\cite{yu2020bdd100k} and OpenDV~\cite{yang2024genad} datasets. As shown in Table~\ref{tab:dataset}, we select a limited number of hours from each dataset and extract frames at 10 Hz. In total, we use 280 hours of video data from a combined available total of 2747 hours. For BDD100K, we select the \textit{day-clear} subset of the training set. From OpenDV we exclude night drives via a brightness filter and uneventful ones by the presence of certain words in the original video titles (see Appendix). We then subsample by selecting evenly spaced 30-second clips.
To train the tokenizer we additionally select images from Honda HAD~\cite{kim2019CVPR}, Honda HDD~\cite{ramanishka2018CVPR}, ONCE~\cite{mao2021one}, NuScenes~\cite{Caesar_2020_CVPR}, and NuPlan~\cite{caesar2021nuplan} to make the dataset diverse. Our dataset primarily consists of daylight scenarios.

\textbf{Tokenizer details.}
For the tokenizer, we employ a Transformer-based encoder and a CNN-based decoder. 
Our tokenizer consists of 234 M parameters and uses two ViT~\cite{dosovitskiy2020image} encoders initialized with MAE~\cite{he2022masked} weights, for the two factorized tokens.
To address codebook under-utilization issues, we incorporate L2-normalized codes~\cite{yu2021vector}, low-latent dimension~\cite{yu2021vector} and entropy penalty~\cite{yu2023language}.
For further improvement, we fine-tune the model with implicit regularization as proposed in EQ-VAE~\cite{kouzelis2025eq}. 
% The tokenizer is trained without the adversarial loss, similar to Cosmos~\cite{agarwal2025cosmos}, and only the decoder is fine-tuned at the end with the adversarial loss for improving the image quality. 

%We refer to our continuous flow matching model as Orbis-FM and the discrete counterpart as Orbis-MG, where FM and MG stand for flow matching and masked generative, respectively.
%We describe the setting used for both world models, the dataset, and 
%perform ablations to study the efficacy of each design choice.

\textbf{Latent world model details.}
Both continuous and discrete models follow a factorized spatial-temporal (ST) transformer architecture~\cite{xu2020spatial}.
For high-resolution experiments, we replace the spatial block with a Swin~\cite{liu2021swin} transformer block for scalability.

% \paragraph{Flow matching model.}
For the FM model, we use DiT~\cite{peebles2023scalable} with ST transformer blocks (STDiT), where a per-frame causal attention mask is applied to the time-attention layers.
%we design a STDiT architecture, a modified version of DiT~\cite{} architecture. 
%As our tokenizer provides two types of embeddings: semantic and detail. 
%We compute the standard deviation of the training set's encoded representations and normalize each frame by dividing by this value, ensuring unit variance across inputs ~\cite{rombach2022high}. 
%This normalization occurs for detail and semantic tokens independently. 
To improve generalization and frame generation quality, we drop all context frames 50\% of the time.
%This number reduces to 10\% after 5 epochs of training. 
When context frames are present, we augment them with noise 50\% of the time, similar to prior work~\cite{valevski2024diffusion, gao2025adaworld, he2022latent}. 
In order to sample the next frame, we use ODE sampler and take 30 steps.
% During training, the context frames are augmented to improve robustness to noisy
% context, particularly during long rollouts of future frames where predicted frames are used as context. 
For the MGM model, we also add context noise to improve robustness towards context noise, especially for long-rollouts:
we replace 10\% of the frames and 10\% of the overall tokens with a mask token.
% We employ 10\% random frame dropping in the context and mask random tokens with 10\% probability.

Masked generative models often exhibit flickering artifacts caused by inconsistent predictions across the temporal dimension. 
We train a lightweight 30M-parameter temporal refinement module to smoothen spatial flickering artifacts. It is U-Net architecture~\cite{ronneberger2015u}, trained using a flow-matching objective. This module operates purely as a post-processing step, on single frames, and does not interfere with the world model. More details are included in the Appendix~\ref{app:sec:model_mgm}.

% We employ two main augmentation strategies: random frame dropping and random token dropping. In both cases, the dropped tokens are replaced with the special mask token.

\textbf{Training details.} Our higher resolution model operates at $512 \times 288$ and small-scale model at $256 \times 256$. Tokenizer compresses the image spatially by $16\times$. We train latent models with a context of 5 frames sampled at 5Hz. 
All small-scale models for ablation studies are trained on only the BDD100K subset for one day on $32\times$ A100 GPUs. The higher resolution model is trained for 10 epochs over 5 days on $72\times$ A100 GPUs.

\subsection{Evaluation}

\subsubsection{Video Generation Quality}
We evaluate the quality of the generated videos using FVD ~\cite{unterthiner2019fvd} and JEDi~\cite{luo2024jedi}, and we use FID to assess tokenizer reconstruction quality. 
For comparing with the baselines, we show FVD and JEDi on nuPlan~\cite{caesar2021nuplan} and Waymo~\cite{sun2020scalability} datasets with 800 and 400 samples respectively. 
To evaluate the models in challenging scenarios, we curate a dedicated validation set of turning events (nuPlan-turns), consisting of 400 samples, selected from the nuPlan validation set where the initial yaw rate is at least 0.12 rad/s ($\sim$1 std). FVD results on nuPlan and nuPlan-turns are not comparable, since nuPlan is a much more diverse dataset compared to the specialized nuPlan-turns dataset.
All evaluation datasets constitute unseen testing domains for our model and for the baselines, except for DrivingWorld, which contains nuPlan as part of the training dataset. 
We choose nuPlan over the similar nuScenes due to the latter's irregular sampling rate, which adds an unnecessary confounding factor to the evaluation.
{Additionally, we evaluate our model using Video Quality Assessment (VQA) metrics including PSNR, SSIM and DOVER\cite{wu2023exploring}, see Appendix~\ref{app:sec:vqa}.}

% \subsubsection{Tokenization Quality}
% We evaluate the tokenizers' ability to reconstruct their inputs using the rFID metric, and the performance of their respective second stage using FVD as described above.
\subsubsection{Trajectory Quality}
\label{sec:traj}

\begin{wraptable}{r}{0.4\textwidth}
\small
  \centering
  \caption{Quality of estimated 10s trajectories for Vista, GEM, and our model, evaluated on turning events from nuPlan.}
  \label{tab:traj-eval}
  \begin{tabular}{lcccc}
    \toprule
    & \multicolumn{2}{c}{Frechet} & \multicolumn{2}{c}{ADE}\\
    Model & Prec. & Rec. & Prec. & Rec. \\
    \midrule
    Vista & 0.39 & 0.45 & 0.25 & 0.48 \\
    GEM   & 0.33 & 0.54 & 0.27 & 0.47 \\
    % DW    & 0.28 & 0.47 & 0.27 & 0.31 \\
    Ours    & \textbf{0.47} & \textbf{0.56} & \textbf{0.41} & \textbf{0.51}\\
    \bottomrule
  \end{tabular}
  \vspace*{0.5cm}
\end{wraptable} 

To evaluate the realism and coverage of generated videos in a manner well suited to driving scenes, where the realism of ego motion and driving behavior is critical, we propose distribution-level, trajectory-based precision and recall metrics. To this end, we map both real and generated videos to pose sequences using the inverse dynamics model VGGT~\cite{wang2025vggt}, and evaluate realism and coverage via precision–recall following~\cite{kynkaanniemi2019pr}, 
% which relies on the number of nearest neighbors within the distribution to determine the distance threshold. 
where the number of nearest neighbors within the distribution determines the distance threshold.
To measure distances over the trajectory sets, we use discrete Fréchet distance~\cite{EiterMannila1994} and Average Displacement Error (ADE)~\cite{pellegrini2009you}, both within the distributions of real and generated trajectories and across them (see Appendix~\ref{app:sec:traj_eval} for full definitions). 
The latter is a stricter metric, as the former is agnostic to velocity differences between trajectories.

We compare the quality of generated trajectories for Vista, GEM, and our approach in (Table~\ref{tab:traj-eval}). The results show the limitations of existing models in capturing the underlying distribution of ego motion and driving behavior. All approaches perform worse in terms of ADE, indicating difficulty in maintaining realistic speeds. Moreover, we achieve the best precision-recall on Fréchet distance, indicating that our predicted trajectories more closely follow the ground‐truth paths compared to competing baselines.

\subsection{Results}\label{sec:res}

\begin{table}
    \centering
    \caption{SOTA results: FVD and JEDi over 6 seconds rollouts@ 5Hz. Numbers of baselines were computed using their official checkpoints. Lower is better. *DrivingWorld (DW) is trained on the test dataset nuPlan and uses ego-motion control as an extra input. Video samples available in Appendix.}
    \vspace{2 mm}
    \label{table:main_results}
    \begin{tabular}{lccc|ccc}
    \toprule
      & \multicolumn{3}{c|}{FVD$\downarrow$} & \multicolumn{3}{c}{JEDi$\downarrow$} \\
      Model  & nuPlan & Waymo & nuPlan & nuPlan & Waymo & nuPlan \\
       &  &  &  turns  &  &  &  turns\\
      \midrule
      Cosmos~\cite{agarwal2025cosmos}  & 291.80 & 278.19 & 248.39   & 0.55 & \textbf{0.31} & 0.50 \\
      Vista~\cite{gao2024vista}        & 323.37 & 422.58 & 413.61   & 0.37 & 0.44 & 0.54\\
      GEM~\cite{hassan2024gem}         & 431.69 & 291.84 & 357.25   & 0.42 & 0.35 & 0.31 \\
      DW*~\cite{hu2024drivingworld}    & 298.97 & N/A & 334.89      & 1.33 & N/A  & 1.50 \\
      \midrule
      % Orbis (ours)  & \textbf{132.25} & \textbf{180.54} & \textbf{231.88}\\ %  199.41
      Orbis (ours)  & \textbf{134.06} & \textbf{167.57} & \textbf{239.20} & \textbf{0.14} & 0.33 & \textbf{0.16} \\ %  199.41
      \bottomrule
    \end{tabular}
\end{table}

\paragraph{Comparison to SOTA.} 
We compare our method against the state-of-the-art video world models for autonomous driving: Vista~\cite{gao2024vista}, GEM~\cite{hassan2024gem}, DrivingWorld (DW)~\cite{hu2024drivingworld}, and the more general-purpose Cosmos~\cite{agarwal2025cosmos} in its autoregressive Predict1-4B version. 
We focus our comparison on steering-free unconditional generation, i.e. with previous visual observations as sole conditioning, with the exception of DrivingWorld which requires the past trajectory. We use a context size of five frames for Vista and DrivingWorld, one for GEM and nine for Cosmos -- as per their respective designs. The input control for DrivingWorld is implemented for nuPlan's data format.

\begin{wrapfigure}{r}{0.45\textwidth}
  \centering
  \includegraphics[width=\linewidth]{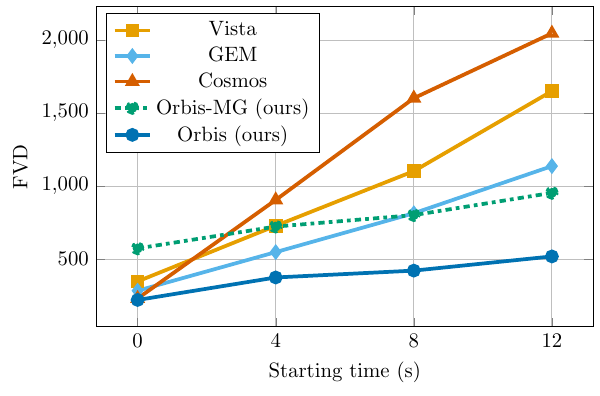}
    \caption{Video quality (FVD) over consecutive 4s time windows on nuPlan-turns. The x axis shows the starting time of the evaluated time window.}
    \label{fig:long_horizon_results}    
\end{wrapfigure}

Results are shown in the Table~\ref{table:main_results}, for a prediction horizon of 6 seconds at 5hz. The qualitative results for 20s are shown in Figure~\ref{fig:qualitaive-rollout} (more are included in Appendix). Our method outperforms other driving world models on all three benchmarks and both metrics, except for Cosmos on Waymo according to the JEDi metric.
We further compare results for long-horizon video prediction, shown in Figure~\ref{fig:long_horizon_results} on nuPlan-turns dataset. For each method, FVD is computed over the entire predicted video in a windowed manner, where each window contains 20 frames sampled at 5hz. Results show that Orbis based on flow matching outperforms all baselines and maintains stable performance over long-horizon prediction of up to 20 seconds. The discrete counterpart Orbis-MG based on MaskGIT, shows suboptimal performance for shorter horizons but scales well over long horizons, surpassing all previous works for the last two windows. As discussed earlier, previous works perform well in short horizons but struggle with long-horizon predictions. GEM has higher FVD scores for short-horizon due to its single-frame 
conditioning design but performs relatively better on long-horizon predictions (more details in the Appendix~\ref{app:sec:more_results}).

\paragraph{Ego-motion Control and Evaluation}
As a proof-of-concept for ego-motion control, we fine-tune a copy of Orbis model for two epochs on 75h of nuPlan videos and IMU data. Following previous literature~\cite{gao2024vista, hassan2024gem, hu2024drivingworld}, we evaluate the resulting model by computing the ADE~\cite{pellegrini2009you} between true and generated trajectories, estimated with VGGT. We compare the ADE of the same model with and without steering on 400 5s long nuPlan validation sequences in Table~\ref{tab:steering}. 
Better trajectory tracking under ego-motion conditioning indicates some degree of controllability -- though in a preliminary setting. Indeed, conditioning capabilities for related models are well documented~\cite{peebles2023scalable,bar2024navigation}.

\begin{table}[!ht]
  \centering
  \begin{minipage}{0.56\textwidth}
    \centering
    \caption{Tokenizer ablation. rFID is computed on 10k BDD100k images. FVD on 200 sequences of 60 frames.}
    \label{table:token_factorization_1}
    \begin{tabular}{lccccc}
      \toprule
      DINO  & TF               & Vocab   & rFID~$\downarrow$      & FVD~$\downarrow$ & FVD~$\downarrow$  \\
        &                &  Size  &     & Orbis-MG& Orbis  \\
      \midrule
      \cross  & \cross        &   4096      & 9.33   &  1331.28  &240.34\\
      \tick   & \cross        &   4096      & 12.17   &  1214.34  &  248.79 \\
      \tick   & \tick         &   2$\times$4096  & 9.10   & 533.28 &  246.11\\
      \bottomrule
    \end{tabular}
  \end{minipage}%
  % \hfill
  \hspace{20pt}  
  \begin{minipage}{0.33\textwidth}
    \centering
    \caption{Ego-motion control: effect on the average error between real and generated trajectories.}
    \label{tab:steering}
    
    \begin{tabular}{cc}
      \toprule
      Model & ADE $\downarrow$ \\
      \midrule
      Unconditional & 5.20\\
      $+$ ego-motion  & 2.40\\
      \bottomrule
      % \vspace{1 mm}
    \end{tabular}
  \end{minipage}
\end{table}

\paragraph{Effect of tokenizer design.}
%To evaluate the importance of the encoded latent representation, we run ablation study and highlight the effects of token factorization on world modeling.

For the discrete model, adding DINO distillation to the image tokenizer, similar to GAIA-1~\cite{hu2023gaia}, leads to lower FVD, as shown in Table~\ref{table:token_factorization_1}. However, the key factor to enable long-horizon prediction for the discrete model is token factorization.
Usage of DINO distillation even leads to a worse rFID (reconstruction FID). However, token factorization annihilates this difference.
Interestingly, while the factorized tokenizer with DINO distillation is very important for the discrete model, the continuous model is robust to these design changes, showing no large change in FVD. These experiments were conducted in the small-scale setting.

\paragraph{Shortcomings of discrete space modeling.}

Despite being capable of relatively long rollouts, the videos produced by Orbis-MG on average stop earlier than its flow matching counterpart, and their duration is very sensitive to the sampling heuristics. 
We investigated this phenomenon and found that at each location, the model's classifier chooses the exact same token as the last context frame approximately 45\% of the times (this number is 29\% for original encoded frames). This is likely because in the discrete space content copying is an obvious and most rewarding choice.
% This strong "zero motion" bias can be explained with the discreteness of the input tokens, which make copying the most rewarding mode. 
While this phenomenon can be mitigated with regularization like context augmentation and a token-similarity based loss, it does not get fully resolved. Additionally, the discrete model fails to capture small motions of objects which is crucial for driving scenarios - thereby limiting the expressivity of the world model.

\begin{wraptable}{r}{0.5\textwidth}
\small
  \centering
  \caption{FVD scores for Orbis-MG and Orbis-FM model with different architectures. The FVD is over 60 generated frames of BDD val set.}
  \label{tab:arch}
  \begin{tabular}{cccc}
    \toprule
         Model & Architecture & 12 steps & 30 steps \\  %& 12 steps & 30 steps\\
        \midrule
        Orbis-MG & ST    &  533.3   &  571.7 \\ %  &  2.32   & 1.07  \\
        (discrete)        & DiT    &  769.0     & 981.3  \\  %   &  2.44   & 1.01  \\
                 & CDiT   &  1552.3     &  1718.9 \\ % ep 7.5 
        \midrule
        Orbis-FM & ST    & 360.9  &   246.0 \\  %&   0.88   &      \\
         (conti.)       & DiT    &  345.4  &  274.2 \\  %&      &      \\
                & CDiT   & 410.2  & 246.1   \\% &  2.76  &   1.09   \\
        \bottomrule
  \end{tabular}
  \vspace*{0.5cm}
\end{wraptable}

\paragraph{Consistency over architectures and sampling budgets.}
To assess the generality of our findings, we evaluate Orbis-FM and MG using three Transformer architectures -- DiT \cite{peebles2023scalable}, STDiT (our default), and CDiT \cite{bar2025navigation}, and with different sampling budgets. All models were trained for one day on 32 A100 GPUs in a small-scale setting, and FVD scores were computed on 200 generated BDD validation videos (60 frames at 5 fps). 
As shown in Table~\ref{tab:arch}, Orbis-FM consistently outperforms Orbis-MG across all architectures and inference settings (12 and 30 steps). Orbis-FM shows greater consistency across architectures than Orbis-MG. The ST architecture shows the best results for both models, with Orbis-FM (ST, 30 steps) achieving the best performance overall.

\begin{figure}
    \centering
    \includegraphics[width=\linewidth]{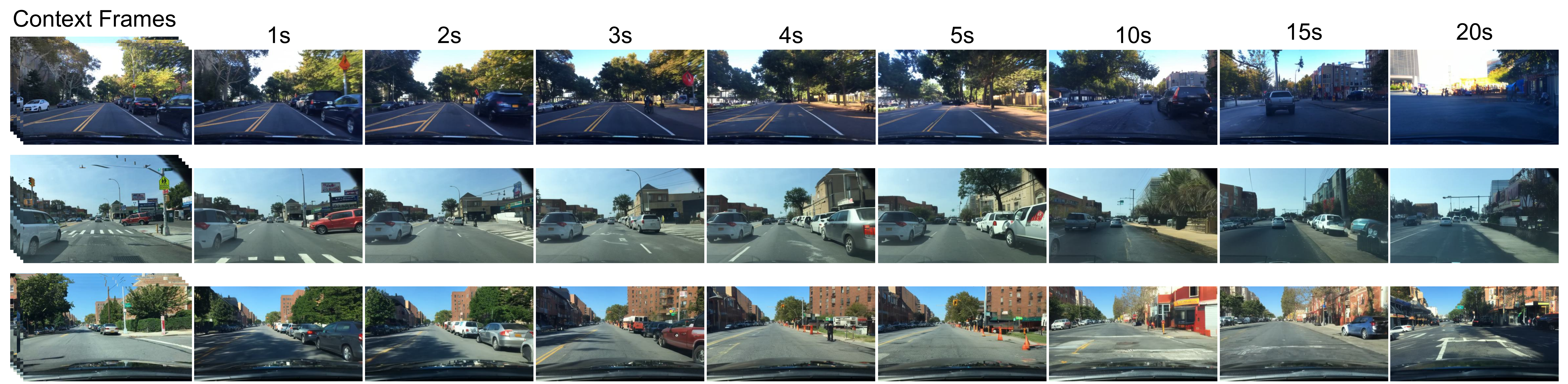}
    \caption{Qualitative results of the Orbis model over 20-sec rollouts (zoom-in for details). Videos and more samples available on the webpage.}
    \label{fig:qualitaive-rollout}
\end{figure}

\section{Discussion}
% 
% We revealed an important shortcoming of contemporary driving world models: their lacking ability for longer roll-outs and turning maneuvers. 
We investigated an important shortcoming of contemporary driving world models: their struggle with the generation of new content, which makes long roll-outs, turning maneuvers, and realistic trajectories impossible.
We introduced an evaluation benchmark and metrics to quantify these problems and tested the hypothesis that modeling in continuous space is the cause of this problem. We found that this is not the case. 
Based on a side-by-side comparison with a fully compatible hybrid tokenizer, we obtained two driving world models that both provide long roll-outs. 
However, the continuous model based on flow matching performs much better and sets the new state of the art. 
The resulting world model has only 469M parameters and was trained on only 280 hours of raw video data. This is significantly less than existing models. At the same time, the approach is perfectly scalable. In contrast to many other recent approaches, it only requires raw video data for training. While we were limited on computing resources for scaling the model ourselves, we expect further improvements when scaling the model parameters, the hours of observed data, the image resolution, and the context length.   

\textbf{Limitations:} While our investigation showed that world models built in continuous space are advantageous over models built in a quantized token space, we were not able to uncover the reason why the much larger public video diffusion models fail on long roll-outs. 
%This could be due to multiple reasons. 
One of the possible reason for this could be that these models are typically (but not always) derived from a pretrained Stable Video Diffusion model. This could introduce biases in the representation, which are problematic for learning relevant state transitions and generating long roll-outs for driving case. %Previous work already showed that off-the-shelf video diffusion models are not good world models~\cite{Worldmodelsandphysics}. 
%Another reason could be the parallel prediction of multiple frames by these models. In contrast, our models -- as well as the GAIA-1 model -- run in a frame-by-frame autoregressive manner. 
%As is known from LLMs, predicting the future content in larger chunks reduces the quality of the generated long-term content. 
%We suspect that this is the main reason for the failure of contemporary video diffusion models in generating long roll-outs, and 
We will analyze this in more detail as future work.   

Apart from this analytic question, our world model has still several limitations, many of which can probably be mitigated by scaling the model along multiple axes. 
Detailed content, such as traffic lights and street signs, are not yet generated reliably. The traffic actors do not always follow the traffic rules. 
While our model shows a good diversity when running multiple roll-outs with the same context, the generated trajectories do not seem to represent the true probability distribution. 
While we showed that conditioning modality can be added flexibly to the model, we did not yet investigate the effectiveness of our model on downstream tasks, such as short-term decision making or planning.  

%In this work, we presented Orbis, a driving world model formulated in both continuous and discrete representation spaces. We conducted targeted studies on the tokenizer design, the dynamics modeling, and the behavior of continuous versus discrete world models. Our results show that the continuous variant of Orbis outperforms prior approaches while using significantly fewer parameters, demonstrating the efficiency of our model. Furthermore, we evaluated the realism of our rollouts by analyzing the generated trajectories. We provide an assessment of Orbis and other driving world models in world driving dynamics.

\textbf{Societal Impact:}
In this work, we contributed to the building of world models -- a technology, which may enable more reliable and cost-efficient autonomous driving and may play a key role in interactive robotics. In its present state, the research is still in its infancy and results that will affect society will still require a few years.

\section*{Acknowledgement}
This work is funded by the German Federal Ministry for Economic Affairs and Energy within the project “NXT GEN AI METHODS" (19A23014R). 
The authors gratefully acknowledge the Gauss Centre for Supercomputing e.V. (www.gauss-centre.eu) for funding this project by providing computing time through the John von Neumann Institute for Computing (NIC) on the GCS Supercomputer JUWELS~\cite{JUWELS} at Jülich Supercomputing Centre (JSC). The project was accomplished under GCS compute grants - \textit{genai-ad} and \textit{nxtaim}.
The compute used in this project is also funded by the German Research Foundation (DFG) - 417962828, 539134284.
The authors acknowledge support from the state of Baden-W\"urttemberg through bwHPC.
We thank Marcel Aach and Sabrina Benassou for their technical support and assistance with the GCS HPC cluster. We thank Rajat Sahay for valuable discussions and feedback during the development of this work.

\newpage

\bibliographystyle{plain}
\bibliography{references}

@inproceedings{gao2024vista,
 title={Vista: A Generalizable Driving World Model with High Fidelity and Versatile Controllability}, 
 author={Shenyuan Gao and Jiazhi Yang and Li Chen and Kashyap Chitta and Yihang Qiu and Andreas Geiger and Jun Zhang and Hongyang Li},
 booktitle={Advances in Neural Information Processing Systems (NeurIPS)},
 year={2024}
}

@inproceedings{yang2024genad,
  title={Generalized Predictive Model for Autonomous Driving},
  author={Jiazhi Yang and Shenyuan Gao and Yihang Qiu and Li Chen and Tianyu Li and Bo Dai and Kashyap Chitta and Penghao Wu and Jia Zeng and Ping Luo and Jun Zhang and Andreas Geiger and Yu Qiao and Hongyang Li},
  booktitle={Proceedings of the IEEE/CVF Conference on Computer Vision and Pattern Recognition (CVPR)},
  year={2024}
}

@InProceedings{Yu_2023_CVPR,
    author    = {Yu, Lijun and Cheng, Yong and Sohn, Kihyuk and Lezama, Jos\'e and Zhang, Han and Chang, Huiwen and Hauptmann, Alexander G. and Yang, Ming-Hsuan and Hao, Yuan and Essa, Irfan and Jiang, Lu},
    title     = {MAGVIT: Masked Generative Video Transformer},
    booktitle = {Proceedings of the IEEE/CVF Conference on Computer Vision and Pattern Recognition (CVPR)},
    month     = {June},
    year      = {2023},
    pages     = {10459-10469}
}

@article{kondratyuk2023videopoet,
  title={Videopoet: A large language model for zero-shot video generation},
  author={Kondratyuk, Dan and Yu, Lijun and Gu, Xiuye and Lezama, Jos{\'e} and Huang, Jonathan and Schindler, Grant and Hornung, Rachel and Birodkar, Vighnesh and Yan, Jimmy and Chiu, Ming-Chang and others},
  journal={arXiv preprint arXiv:2312.14125},
  year={2023}
}

@article{liu2024mardini,
  title={Mardini: Masked autoregressive diffusion for video generation at scale},
  author={Liu, Haozhe and Liu, Shikun and Zhou, Zijian and Xu, Mengmeng and Xie, Yanping and Han, Xiao and P{\'e}rez, Juan C and Liu, Ding and Kahatapitiya, Kumara and Jia, Menglin and others},
  journal={arXiv preprint arXiv:2410.20280},
  year={2024}
}

@inproceedings{wang2024drivedreamer,
  title={DriveDreamer: Towards Real-World-Drive World Models for Autonomous Driving},
  author={Wang, Xiaofeng and Zhu, Zheng and Huang, Guan and Chen, Xinze and Zhu, Jiagang and Lu, Jiwen},
  booktitle={European Conference on Computer Vision},
  pages={55--72},
  year={2024},
  organization={Springer}
}

@inproceedings{wang2024driving,
  title={Driving into the future: Multiview visual forecasting and planning with world model for autonomous driving},
  author={Wang, Yuqi and He, Jiawei and Fan, Lue and Li, Hongxin and Chen, Yuntao and Zhang, Zhaoxiang},
  booktitle={Proceedings of the IEEE/CVF Conference on Computer Vision and Pattern Recognition},
  pages={14749--14759},
  year={2024}
}

@inproceedings{kim2021drivegan,
  title={Drivegan: Towards a controllable high-quality neural simulation},
  author={Kim, Seung Wook and Philion, Jonah and Torralba, Antonio and Fidler, Sanja},
  booktitle={Proceedings of the IEEE/CVF Conference on Computer Vision and Pattern Recognition},
  pages={5820--5829},
  year={2021}
}

@article{hu2023gaia,
  title={Gaia-1: A generative world model for autonomous driving},
  author={Hu, Anthony and Russell, Lloyd and Yeo, Hudson and Murez, Zak and Fedoseev, George and Kendall, Alex and Shotton, Jamie and Corrado, Gianluca},
  journal={arXiv preprint arXiv:2309.17080},
  year={2023}
}

@article{hassan2024gem,
  title={GEM: A Generalizable Ego-Vision Multimodal World Model for Fine-Grained Ego-Motion, Object Dynamics, and Scene Composition Control},
  author={Hassan, Mariam and Stapf, Sebastian and Rahimi, Ahmad and Rezende, Pedro and Haghighi, Yasaman and Br{\"u}ggemann, David and Katircioglu, Isinsu and Zhang, Lin and Chen, Xiaoran and Saha, Suman and others},
  journal={arXiv preprint arXiv:2412.11198},
  year={2024}
}

@misc{oquab2023dinov2,
  title={DINOv2: Learning Robust Visual Features without Supervision},
  author={Oquab, Maxime and Darcet, Timothée and Moutakanni, Theo and Vo, Huy V. and Szafraniec, Marc and Khalidov, Vasil and Fernandez, Pierre and Haziza, Daniel and Massa, Francisco and El-Nouby, Alaaeldin and Howes, Russell and Huang, Po-Yao and Xu, Hu and Sharma, Vasu and Li, Shang-Wen and Galuba, Wojciech and Rabbat, Mike and Assran, Mido and Ballas, Nicolas and Synnaeve, Gabriel and Misra, Ishan and Jegou, Herve and Mairal, Julien and Labatut, Patrick and Joulin, Armand and Bojanowski, Piotr},
  journal={arXiv:2304.07193},
  year={2023}
}

@inproceedings{bruce2024genie,
  title={Genie: Generative interactive environments},
  author={Bruce, Jake and Dennis, Michael D and Edwards, Ashley and Parker-Holder, Jack and Shi, Yuge and Hughes, Edward and Lai, Matthew and Mavalankar, Aditi and Steigerwald, Richie and Apps, Chris and others},
  booktitle={Forty-first International Conference on Machine Learning},
  year={2024}
}

@inproceedings{chang2022maskgit,
  title={Maskgit: Masked generative image transformer},
  author={Chang, Huiwen and Zhang, Han and Jiang, Lu and Liu, Ce and Freeman, William T},
  booktitle={Proceedings of the IEEE/CVF conference on computer vision and pattern recognition},
  pages={11315--11325},
  year={2022}
}

@article{bai2024factorized,
  title={Factorized Visual Tokenization and Generation},
  author={Bai, Zechen and Gao, Jianxiong and Gao, Ziteng and Wang, Pichao and Zhang, Zheng and He, Tong and Shou, Mike Zheng},
  journal={arXiv preprint arXiv:2411.16681},
  year={2024}
}

@inproceedings{
li2025imagefolder,
title={ImageFolder: Autoregressive Image Generation with Folded Tokens},
author={Xiang Li and Kai Qiu and Hao Chen and Jason Kuen and Jiuxiang Gu and Bhiksha Raj and Zhe Lin},
booktitle={The Thirteenth International Conference on Learning Representations},
year={2025},
url={https://openreview.net/forum?id=QE1LFzXQPL}
}

@article{qu2024tokenflow,
  title={Tokenflow: Unified image tokenizer for multimodal understanding and generation},
  author={Qu, Liao and Zhang, Huichao and Liu, Yiheng and Wang, Xu and Jiang, Yi and Gao, Yiming and Ye, Hu and Du, Daniel K and Yuan, Zehuan and Wu, Xinglong},
  journal={arXiv preprint arXiv:2412.03069},
  year={2024}
}

@article{agarwal2025cosmos,
  title={Cosmos world foundation model platform for physical ai},
  author={Agarwal, Niket and Ali, Arslan and Bala, Maciej and Balaji, Yogesh and Barker, Erik and Cai, Tiffany and Chattopadhyay, Prithvijit and Chen, Yongxin and Cui, Yin and Ding, Yifan and others},
  journal={arXiv preprint arXiv:2501.03575},
  year={2025}
}

@article{blattmann2023stable,
  title={Stable video diffusion: Scaling latent video diffusion models to large datasets},
  author={Blattmann, Andreas and Dockhorn, Tim and Kulal, Sumith and Mendelevitch, Daniel and Kilian, Maciej and Lorenz, Dominik and Levi, Yam and English, Zion and Voleti, Vikram and Letts, Adam and others},
  journal={arXiv preprint arXiv:2311.15127},
  year={2023}
}

@article{van2017neural,
  title={Neural discrete representation learning},
  author={Van Den Oord, Aaron and Vinyals, Oriol and others},
  journal={Advances in neural information processing systems},
  volume={30},
  year={2017}
}

@article{yu2023language,
  title={Language Model Beats Diffusion--Tokenizer is Key to Visual Generation},
  author={Yu, Lijun and Lezama, Jos{\'e} and Gundavarapu, Nitesh B and Versari, Luca and Sohn, Kihyuk and Minnen, David and Cheng, Yong and Birodkar, Vighnesh and Gupta, Agrim and Gu, Xiuye and others},
  journal={arXiv preprint arXiv:2310.05737},
  year={2023}
}

@article{kouzelis2025eq,
  title={EQ-VAE: Equivariance Regularized Latent Space for Improved Generative Image Modeling},
  author={Kouzelis, Theodoros and Kakogeorgiou, Ioannis and Gidaris, Spyros and Komodakis, Nikos},
  journal={arXiv preprint arXiv:2502.09509},
  year={2025}
}

@article{yu2021vector,
  title={Vector-quantized image modeling with improved vqgan},
  author={Yu, Jiahui and Li, Xin and Koh, Jing Yu and Zhang, Han and Pang, Ruoming and Qin, James and Ku, Alexander and Xu, Yuanzhong and Baldridge, Jason and Wu, Yonghui},
  journal={arXiv preprint arXiv:2110.04627},
  year={2021}
}

@inproceedings{esser2021taming,
  title={Taming transformers for high-resolution image synthesis},
  author={Esser, Patrick and Rombach, Robin and Ommer, Bjorn},
  booktitle={Proceedings of the IEEE/CVF conference on computer vision and pattern recognition},
  pages={12873--12883},
  year={2021}
}

@article{lipman2022flow,
  title={Flow matching for generative modeling},
  author={Lipman, Yaron and Chen, Ricky TQ and Ben-Hamu, Heli and Nickel, Maximilian and Le, Matt},
  journal={arXiv preprint arXiv:2210.02747},
  year={2022}
}

@article{hafner2019dream,
  title={Dream to control: Learning behaviors by latent imagination},
  author={Hafner, Danijar and Lillicrap, Timothy and Ba, Jimmy and Norouzi, Mohammad},
  journal={arXiv preprint arXiv:1912.01603},
  year={2019}
}

@inproceedings{piergiovanni2019learning,
  title={Learning real-world robot policies by dreaming},
  author={Piergiovanni, AJ and Wu, Alan and Ryoo, Michael S},
  booktitle={2019 IEEE/RSJ International Conference on Intelligent Robots and Systems (IROS)},
  pages={7680--7687},
  year={2019},
  organization={IEEE}
}

@article{wu2023pre,
  title={Pre-training contextualized world models with in-the-wild videos for reinforcement learning},
  author={Wu, Jialong and Ma, Haoyu and Deng, Chaoyi and Long, Mingsheng},
  journal={Advances in Neural Information Processing Systems},
  volume={36},
  pages={39719--39743},
  year={2023}
}

@article{he2024large,
  title={Large-scale actionless video pre-training via discrete diffusion for efficient policy learning},
  author={He, Haoran and Bai, Chenjia and Pan, Ling and Zhang, Weinan and Zhao, Bin and Li, Xuelong},
  journal={arXiv e-prints},
  pages={arXiv--2402},
  year={2024}
}

@article{zhao2024drivedreamer4d,
  title={Drivedreamer4d: World models are effective data machines for 4d driving scene representation},
  author={Zhao, Guosheng and Ni, Chaojun and Wang, Xiaofeng and Zhu, Zheng and Zhang, Xueyang and Wang, Yida and Huang, Guan and Chen, Xinze and Wang, Boyuan and Zhang, Youyi and others},
  journal={arXiv preprint arXiv:2410.13571},
  year={2024}
}

@article{yang2024cogvideox,
  title={Cogvideox: Text-to-video diffusion models with an expert transformer},
  author={Yang, Zhuoyi and Teng, Jiayan and Zheng, Wendi and Ding, Ming and Huang, Shiyu and Xu, Jiazheng and Yang, Yuanming and Hong, Wenyi and Zhang, Xiaohan and Feng, Guanyu and others},
  journal={arXiv preprint arXiv:2408.06072},
  year={2024}
}

@inproceedings{li2024think2drive,
  title={Think2Drive: Efficient Reinforcement Learning by Thinking with Latent World Model for Autonomous Driving (in CARLA-V2)},
  author={Li, Qifeng and Jia, Xiaosong and Wang, Shaobo and Yan, Junchi},
  booktitle={European Conference on Computer Vision},
  pages={142--158},
  year={2024},
  organization={Springer}
}

@inproceedings{liu2021swin,
  title={Swin transformer: Hierarchical vision transformer using shifted windows},
  author={Liu, Ze and Lin, Yutong and Cao, Yue and Hu, Han and Wei, Yixuan and Zhang, Zheng and Lin, Stephen and Guo, Baining},
  booktitle={Proceedings of the IEEE/CVF international conference on computer vision},
  pages={10012--10022},
  year={2021}
}

@article{valevski2024diffusion,
  title={Diffusion models are real-time game engines},
  author={Valevski, Dani and Leviathan, Yaniv and Arar, Moab and Fruchter, Shlomi},
  journal={arXiv preprint arXiv:2408.14837},
  year={2024}
}

@article{gao2025adaworld,
  title={Adaworld: Learning adaptable world models with latent actions},
  author={Gao, Shenyuan and Zhou, Siyuan and Du, Yilun and Zhang, Jun and Gan, Chuang},
  journal={arXiv preprint arXiv:2503.18938},
  year={2025}
}

@article{he2022latent,
  title={Latent video diffusion models for high-fidelity long video generation},
  author={He, Yingqing and Yang, Tianyu and Zhang, Yong and Shan, Ying and Chen, Qifeng},
  journal={arXiv preprint arXiv:2211.13221},
  year={2022}
}

@article{zhao2024drive,
  title={DriveDreamer-2: LLM-Enhanced World Models for Diverse Driving Video Generation},
  author={Zhao, Guosheng and Wang, Xiaofeng and Zhu, Zheng and Chen, Xinze and Huang, Guan and Bao, Xiaoyi and Wang, Xingang},
  journal={arXiv preprint arXiv:2403.06845},
  year={2024}
}

@article{jia2023adriver,
  title={Adriver-i: A general world model for autonomous driving},
  author={Jia, Fan and Mao, Weixin and Liu, Yingfei and Zhao, Yucheng and Wen, Yuqing and Zhang, Chi and Zhang, Xiangyu and Wang, Tiancai},
  journal={arXiv preprint arXiv:2311.13549},
  year={2023}
}

@InProceedings{Wen_2024_CVPR,
    author    = {Wen, Yuqing and Zhao, Yucheng and Liu, Yingfei and Jia, Fan and Wang, Yanhui and Luo, Chong and Zhang, Chi and Wang, Tiancai and Sun, Xiaoyan and Zhang, Xiangyu},
    title     = {Panacea: Panoramic and Controllable Video Generation for Autonomous Driving},
    booktitle = {Proceedings of the IEEE/CVF Conference on Computer Vision and Pattern Recognition (CVPR)},
    month     = {June},
    year      = {2024},
    pages     = {6902-6912}
}

@article{xu2020spatial,
  title={Spatial-temporal transformer networks for traffic flow forecasting},
  author={Xu, Mingxing and Dai, Wenrui and Liu, Chunmiao and Gao, Xing and Lin, Weiyao and Qi, Guo-Jun and Xiong, Hongkai},
  journal={arXiv preprint arXiv:2001.02908},
  year={2020}
}

@article{gupta2022maskvit,
  title={Maskvit: Masked visual pre-training for video prediction},
  author={Gupta, Agrim and Tian, Stephen and Zhang, Yunzhi and Wu, Jiajun and Mart{\'\i}n-Mart{\'\i}n, Roberto and Fei-Fei, Li},
  journal={arXiv preprint arXiv:2206.11894},
  year={2022}
}

@inproceedings{10.5555/3692070.3693075,
author = {Kondratyuk, Dan and Yu, Lijun and Gu, Xiuye and Lezama, Jos\'{e} and Huang, Jonathan and Schindler, Grant and Hornung, Rachel and Birodkar, Vighnesh and Yan, Jimmy and Chiu, Ming-Chang and Somandepalli, Krishna and Akbari, Hassan and Alon, Yair and Cheng, Yong and Dillon, Josh and Gupta, Agrim and Hahn, Meera and Hauth, Anja and Hendon, David and Martinez, Alonso and Minnen, David and Sirotenko, Mikhail and Sohn, Kihyuk and Yang, Xuan and Adam, Hartwig and Yang, Ming-Hsuan and Essa, Irfan and Wang, Huisheng and Ross, David A. and Seybold, Bryan and Jiang, Lu},
title = {VideoPoet: a large language model for zero-shot video generation},
year = {2024},
publisher = {JMLR.org},
booktitle = {Proceedings of the 41st International Conference on Machine Learning},
articleno = {1005},
numpages = {20},
location = {Vienna, Austria},
series = {ICML'24}
}

@article{yan2021videogpt,
  title={Videogpt: Video generation using vq-vae and transformers},
  author={Yan, Wilson and Zhang, Yunzhi and Abbeel, Pieter and Srinivas, Aravind},
  journal={arXiv preprint arXiv:2104.10157},
  year={2021}
}

@article{hafner2020mastering,
  title={Mastering atari with discrete world models},
  author={Hafner, Danijar and Lillicrap, Timothy and Norouzi, Mohammad and Ba, Jimmy},
  journal={arXiv preprint arXiv:2010.02193},
  year={2020}
}

@inproceedings{dosovitskiy2017carla,
  title={CARLA: An open urban driving simulator},
  author={Dosovitskiy, Alexey and Ros, German and Codevilla, Felipe and Lopez, Antonio and Koltun, Vladlen},
  booktitle={Conference on robot learning},
  pages={1--16},
  year={2017},
  organization={PMLR}
}

@article{ha2018recurrent,
  title={Recurrent world models facilitate policy evolution},
  author={Ha, David and Schmidhuber, J{\"u}rgen},
  journal={Advances in neural information processing systems},
  volume={31},
  year={2018}
}

@inproceedings{lee2022autoregressive,
  title={Autoregressive image generation using residual quantization},
  author={Lee, Doyup and Kim, Chiheon and Kim, Saehoon and Cho, Minsu and Han, Wook-Shin},
  booktitle={Proceedings of the IEEE/CVF Conference on Computer Vision and Pattern Recognition},
  pages={11523--11532},
  year={2022}
}

@article{tian2024visual,
  title={Visual autoregressive modeling: Scalable image generation via next-scale prediction},
  author={Tian, Keyu and Jiang, Yi and Yuan, Zehuan and Peng, Bingyue and Wang, Liwei},
  journal={Advances in neural information processing systems},
  volume={37},
  pages={84839--84865},
  year={2024}
}

@article{lin2024open,
  title={Open-sora plan: Open-source large video generation model},
  author={Lin, Bin and Ge, Yunyang and Cheng, Xinhua and Li, Zongjian and Zhu, Bin and Wang, Shaodong and He, Xianyi and Ye, Yang and Yuan, Shenghai and Chen, Liuhan and others},
  journal={arXiv preprint arXiv:2412.00131},
  year={2024}
}

@article{esteves2024spectral,
  title={Spectral Image Tokenizer},
  author={Esteves, Carlos and Suhail, Mohammed and Makadia, Ameesh},
  journal={arXiv preprint arXiv:2412.09607},
  year={2024}
}

@incollection{pinheiro2021variational,
  title={Variational autoencoder},
  author={Pinheiro Cinelli, Lucas and Ara{\'u}jo Marins, Matheus and Barros da Silva, Eduardo Ant{\'u}nio and Lima Netto, S{\'e}rgio},
  booktitle={Variational methods for machine learning with applications to deep networks},
  pages={111--149},
  year={2021},
  publisher={Springer}
}

@article{russell2025gaia,
  title={Gaia-2: A controllable multi-view generative world model for autonomous driving},
  author={Russell, Lloyd and Hu, Anthony and Bertoni, Lorenzo and Fedoseev, George and Shotton, Jamie and Arani, Elahe and Corrado, Gianluca},
  journal={arXiv preprint arXiv:2503.20523},
  year={2025}
}

@article{polyak2024movie,
  title={Movie Gen: A Cast of Media Foundation Models},
  author={Polyak, Adam and Zohar, Amit and Brown, Andrew and Tjandra, Andros and Sinha, Animesh and Lee, Ann and Vyas, Apoorv and Shi, Bowen and Ma, Chih-Yao and Chuang, Ching-Yao and others},
  journal={arXiv preprint arXiv:2410.13720},
  year={2024}
}

@article{jin2024pyramidal,
  title={Pyramidal flow matching for efficient video generative modeling},
  author={Jin, Yang and Sun, Zhicheng and Li, Ningyuan and Xu, Kun and Jiang, Hao and Zhuang, Nan and Huang, Quzhe and Song, Yang and Mu, Yadong and Lin, Zhouchen},
  journal={arXiv preprint arXiv:2410.05954},
  year={2024}
}

@article{zhou2022magicvideo,
  title={Magicvideo: Efficient video generation with latent diffusion models},
  author={Zhou, Daquan and Wang, Weimin and Yan, Hanshu and Lv, Weiwei and Zhu, Yizhe and Feng, Jiashi},
  journal={arXiv preprint arXiv:2211.11018},
  year={2022}
}

@article{ma2024star,
  title={Star: Scale-wise text-to-image generation via auto-regressive representations},
  author={Ma, Xiaoxiao and Zhou, Mohan and Liang, Tao and Bai, Yalong and Zhao, Tiejun and Chen, Huaian and Jin, Yi},
  journal={arXiv preprint arXiv:2406.10797},
  year={2024}
}

@article{li2024bevformer,
  title={Bevformer: learning bird's-eye-view representation from lidar-camera via spatiotemporal transformers},
  author={Li, Zhiqi and Wang, Wenhai and Li, Hongyang and Xie, Enze and Sima, Chonghao and Lu, Tong and Yu, Qiao and Dai, Jifeng},
  journal={IEEE Transactions on Pattern Analysis and Machine Intelligence},
  year={2024},
  publisher={IEEE}
}

@article{liu2023visual,
  title={Visual instruction tuning},
  author={Liu, Haotian and Li, Chunyuan and Wu, Qingyang and Lee, Yong Jae},
  journal={Advances in neural information processing systems},
  volume={36},
  pages={34892--34916},
  year={2023}
}

@inproceedings{rombach2022high,
  title={High-resolution image synthesis with latent diffusion models},
  author={Rombach, Robin and Blattmann, Andreas and Lorenz, Dominik and Esser, Patrick and Ommer, Bj{\"o}rn},
  booktitle={Proceedings of the IEEE/CVF conference on computer vision and pattern recognition},
  pages={10684--10695},
  year={2022}
}

@inproceedings{
tang2025hart,
title={{HART}: Efficient Visual Generation with Hybrid Autoregressive Transformer},
author={Haotian Tang and Yecheng Wu and Shang Yang and Enze Xie and Junsong Chen and Junyu Chen and Zhuoyang Zhang and Han Cai and Yao Lu and Song Han},
booktitle={The Thirteenth International Conference on Learning Representations},
year={2025},
url={https://openreview.net/forum?id=q5sOv4xQe4}
}

@inproceedings{peebles2023scalable,
  title={Scalable diffusion models with transformers},
  author={Peebles, William and Xie, Saining},
  booktitle={Proceedings of the IEEE/CVF international conference on computer vision},
  pages={4195--4205},
  year={2023}
}

@article{loshchilov2017decoupled,
  title={Decoupled weight decay regularization},
  author={Loshchilov, Ilya and Hutter, Frank},
  journal={arXiv preprint arXiv:1711.05101},
  year={2017}
}

@misc{
unterthiner2019fvd,
title={{FVD}: A new Metric for Video Generation},
author={Thomas Unterthiner and Sjoerd van Steenkiste and Karol Kurach and Rapha{\"e}l Marinier and Marcin Michalski and Sylvain Gelly},
year={2019},
url={https://openreview.net/forum?id=rylgEULtdN}
}

@misc{luo2024jedi,
        title={Beyond FVD: Enhanced Evaluation Metrics for Video Generation Quality}, 
        author={Ge Ya Luo and Gian Favero and Zhi Hao Luo and Alexia Jolicoeur-Martineau and Christopher Pal},
        year={2024},
        eprint={2410.05203},
        archivePrefix={arXiv},
        primaryClass={cs.CV},
        url={https://arxiv.org/abs/2410.05203}
      }

@article{hu2024drivingworld,
  title={DrivingWorld: ConstructingWorld Model for Autonomous Driving via Video GPT},
  author={Hu, Xiaotao and Yin, Wei and Jia, Mingkai and Deng, Junyuan and Guo, Xiaoyang and Zhang, Qian and Long, Xiaoxiao and Tan, Ping},
  journal={arXiv preprint arXiv:2412.19505},
  year={2024}
}

@inproceedings{yu2020bdd100k,
  title={Bdd100k: A diverse driving dataset for heterogeneous multitask learning},
  author={Yu, Fisher and Chen, Haofeng and Wang, Xin and Xian, Wenqi and Chen, Yingying and Liu, Fangchen and Madhavan, Vashisht and Darrell, Trevor},
  booktitle={Proceedings of the IEEE/CVF conference on computer vision and pattern recognition},
  pages={2636--2645},
  year={2020}
}

@inproceedings{perez2018film,
  title={Film: Visual reasoning with a general conditioning layer},
  author={Perez, Ethan and Strub, Florian and De Vries, Harm and Dumoulin, Vincent and Courville, Aaron},
  booktitle={Proceedings of the AAAI conference on artificial intelligence},
  volume={32},
  year={2018}
}

@inproceedings{he2022masked,
  title={Masked autoencoders are scalable vision learners},
  author={He, Kaiming and Chen, Xinlei and Xie, Saining and Li, Yanghao and Doll{\'a}r, Piotr and Girshick, Ross},
  booktitle={Proceedings of the IEEE/CVF conference on computer vision and pattern recognition},
  pages={16000--16009},
  year={2022}
}

@article{savov2025exploration,
  title={Exploration-Driven Generative Interactive Environments},
  author={Savov, Nedko and Kazemi, Naser and Mahdi, Mohammad and Paudel, Danda Pani and Wang, Xi and Van Gool, Luc},
  journal={arXiv preprint arXiv:2504.02515},
  year={2025}
}

@inproceedings{pellegrini2009you,
  title={You'll never walk alone: Modeling social behavior for multi-target tracking},
  author={Pellegrini, Stefano and Ess, Andreas and Schindler, Konrad and Van Gool, Luc},
  booktitle={2009 IEEE 12th international conference on computer vision},
  pages={261--268},
  year={2009},
  organization={IEEE}
}

@InProceedings{kim2019CVPR,
        author = {Jinkyu Kim and Teruhisa Misu and Yi-Ting Chen and Ashish Tawari and John Canny},
        title = {Grounding Human-To-Vehicle Advice for Self-Driving Vehicles},
        booktitle = {The IEEE Conference on Computer Vision and Pattern Recognition (CVPR)},
        year = {2019}
}

@article{mao2021one,
  title={One million scenes for autonomous driving: Once dataset},
  author={Mao, Jiageng and Niu, Minzhe and Jiang, Chenhan and Liang, Hanxue and Chen, Jingheng and Liang, Xiaodan and Li, Yamin and Ye, Chaoqiang and Zhang, Wei and Li, Zhenguo and others},
  journal={arXiv preprint arXiv:2106.11037},
  year={2021}
}

@inproceedings{ramanishka2018CVPR,
    author    = {Vasili Ramanishka and Yi-Ting Chen and Teruhisa Misu and Kate Saenko},
    title     = {Toward Driving Scene Understanding: A Dataset for Learning Driver Behavior and Causal Reasoning},
    booktitle = {Conference on Computer Vision and Pattern Recognition (CVPR)},
    year      = {2018}
}

@article{caesar2021nuplan,
  title={nuplan: A closed-loop ml-based planning benchmark for autonomous vehicles},
  author={Caesar, Holger and Kabzan, Juraj and Tan, Kok Seang and Fong, Whye Kit and Wolff, Eric and Lang, Alex and Fletcher, Luke and Beijbom, Oscar and Omari, Sammy},
  journal={arXiv preprint arXiv:2106.11810},
  year={2021}
}

@InProceedings{Caesar_2020_CVPR,
author = {Caesar, Holger and Bankiti, Varun and Lang, Alex H. and Vora, Sourabh and Liong, Venice Erin and Xu, Qiang and Krishnan, Anush and Pan, Yu and Baldan, Giancarlo and Beijbom, Oscar},
title = {nuScenes: A Multimodal Dataset for Autonomous Driving},
booktitle = {Proceedings of the IEEE/CVF Conference on Computer Vision and Pattern Recognition (CVPR)},
month = {June},
year = {2020}
}

@inproceedings{sun2020scalability,
  title={Scalability in perception for autonomous driving: Waymo open dataset},
  author={Sun, Pei and Kretzschmar, Henrik and Dotiwalla, Xerxes and Chouard, Aurelien and Patnaik, Vijaysai and Tsui, Paul and Guo, James and Zhou, Yin and Chai, Yuning and Caine, Benjamin and others},
  booktitle={Proceedings of the IEEE/CVF conference on computer vision and pattern recognition},
  pages={2446--2454},
  year={2020}
}

@inproceedings{kynkaanniemi2019pr,
  title     = {Improved precision and recall metric for assessing generative models},
  author    = {Kynk{\"a}{\"a}nniemi, Tuomas and Karras, Tero and Laine, Samuli and Lehtinen, Jaakko and Aila, Timo},
  booktitle = {Advances in Neural Information Processing Systems},
  volume    = {32},
  pages     = {3929--3938},
  year      = {2019}
}

@techreport{EiterMannila1994,
  author       = {Thomas Eiter and Heikki Mannila},
  title        = {Computing Discrete Fr{\'e}chet Distance},
  institution  = {Christian Doppler Laboratory for Expert Systems, TU Vienna},
  year         = {1994},
  number       = {CD-TR 94/64},
  url          = {https://www.kr.tuwien.ac.at/staff/eiter/et-archive/files/cdtr9464.pdf}
}

@article{wang2025vggt,
  title={Vggt: Visual geometry grounded transformer},
  author={Wang, Jianyuan and Chen, Minghao and Karaev, Nikita and Vedaldi, Andrea and Rupprecht, Christian and Novotny, David},
  journal={arXiv preprint arXiv:2503.11651},
  year={2025}
}

@article{ha2018world,
  title={World models},
  author={Ha, David and Schmidhuber, J{\"u}rgen},
  journal={arXiv preprint arXiv:1803.10122},
  year={2018}
}

@article{dosovitskiy2020image,
  title={An image is worth 16x16 words: Transformers for image recognition at scale},
  author={Dosovitskiy, Alexey and Beyer, Lucas and Kolesnikov, Alexander and Weissenborn, Dirk and Zhai, Xiaohua and Unterthiner, Thomas and Dehghani, Mostafa and Minderer, Matthias and Heigold, Georg and Gelly, Sylvain and others},
  journal={arXiv preprint arXiv:2010.11929},
  year={2020}
}

@inproceedings{yu2023magvit,
  title={Magvit: Masked generative video transformer},
  author={Yu, Lijun and Cheng, Yong and Sohn, Kihyuk and Lezama, Jos{\'e} and Zhang, Han and Chang, Huiwen and Hauptmann, Alexander G and Yang, Ming-Hsuan and Hao, Yuan and Essa, Irfan and others},
  booktitle={Proceedings of the IEEE/CVF Conference on Computer Vision and Pattern Recognition},
  pages={10459--10469},
  year={2023}
}

@article{liu2022flow,
  title={Flow straight and fast: Learning to generate and transfer data with rectified flow},
  author={Liu, Xingchao and Gong, Chengyue and Liu, Qiang},
  journal={arXiv preprint arXiv:2209.03003},
  year={2022}
}

@inproceedings{esser2024scaling,
  title={Scaling rectified flow transformers for high-resolution image synthesis},
  author={Esser, Patrick and Kulal, Sumith and Blattmann, Andreas and Entezari, Rahim and M{\"u}ller, Jonas and Saini, Harry and Levi, Yam and Lorenz, Dominik and Sauer, Axel and Boesel, Frederic and others},
  booktitle={Forty-first international conference on machine learning},
  year={2024}
}

@article{bar2024navigation,
  title={Navigation world models},
  author={Bar, Amir and Zhou, Gaoyue and Tran, Danny and Darrell, Trevor and LeCun, Yann},
  journal={arXiv preprint arXiv:2412.03572},
  year={2024}
}

@inproceedings{girdhar2024factorizing,
  title={Factorizing text-to-video generation by explicit image conditioning},
  author={Girdhar, Rohit and Singh, Mannat and Brown, Andrew and Duval, Quentin and Azadi, Samaneh and Rambhatla, Sai Saketh and Shah, Akbar and Yin, Xi and Parikh, Devi and Misra, Ishan},
  booktitle={European Conference on Computer Vision},
  pages={205--224},
  year={2024},
  organization={Springer}
}

@inproceedings{ronneberger2015u,
  title={U-net: Convolutional networks for biomedical image segmentation},
  author={Ronneberger, Olaf and Fischer, Philipp and Brox, Thomas},
  booktitle={Medical image computing and computer-assisted intervention--MICCAI 2015: 18th international conference, Munich, Germany, October 5-9, 2015, proceedings, part III 18},
  pages={234--241},
  year={2015},
  organization={Springer}
}

@article{hinton2015distilling,
  title={Distilling the knowledge in a neural network},
  author={Hinton, Geoffrey and Vinyals, Oriol and Dean, Jeff},
  journal={arXiv preprint arXiv:1503.02531},
  year={2015}
}

@article{JUWELS,
author = {{J\"{u}lich Supercomputing Centre}},
title = {{JUWELS Cluster and Booster: Exascale Pathfinder with Modular Supercomputing Architecture at Juelich Supercomputing Centre}},
journal = {Journal of large-scale research facilities},
number = {A138},
volume = {7},
doi = {10.17815/jlsrf-7-183},
url = {http://dx.doi.org/10.17815/jlsrf-7-183},
year = {2021}
}

@inproceedings{wu2023exploring,
  title={Exploring video quality assessment on user generated contents from aesthetic and technical perspectives},
  author={Wu, Haoning and Zhang, Erli and Liao, Liang and Chen, Chaofeng and Hou, Jingwen and Wang, Annan and Sun, Wenxiu and Yan, Qiong and Lin, Weisi},
  booktitle={Proceedings of the IEEE/CVF International Conference on Computer Vision},
  pages={20144--20154},
  year={2023}
}

@article{assran2025v,
  title={V-jepa 2: Self-supervised video models enable understanding, prediction and planning},
  author={Assran, Mido and Bardes, Adrien and Fan, David and Garrido, Quentin and Howes, Russell and Muckley, Matthew and Rizvi, Ammar and Roberts, Claire and Sinha, Koustuv and Zholus, Artem and others},
  journal={arXiv preprint arXiv:2506.09985},
  year={2025}
}

@inproceedings{bar2025navigation,
  title={Navigation world models},
  author={Bar, Amir and Zhou, Gaoyue and Tran, Danny and Darrell, Trevor and LeCun, Yann},
  booktitle={Proceedings of the Computer Vision and Pattern Recognition Conference},
  pages={15791--15801},
  year={2025}
}

%%%%%%%%%%%%%%%%%%%%%%%%%%%%%%%%%%%%%%%%%%%%%%%%%%%%%%%%%%%%

% \appendix

% \section{Technical Appendices and Supplementary Material}
% Technical appendices with additional results, figures, graphs and proofs may be submitted with the paper submission before the full submission deadline (see above), or as a separate PDF in the ZIP file below before the supplementary material deadline. There is no page limit for the technical appendices.

%%%%%%%%%%%%%%%%%%%%%%%%%%%%%%%%%%%%%%%%%%%%%%%%%%%%%%%%%%%%

\newpage
\section*{NeurIPS Paper Checklist}

\begin{enumerate}
%\answerYes{}, \answerNo{}, or \answerNA{}.
\item {\bf Claims}
    \item[] Question: Do the main claims made in the abstract and introduction accurately reflect the paper's contributions and scope?
    \item[] Answer:  \answerYes{}
    \item[] Justification: The abstract summarizes the problem tackled in this paper and our contributions. The introduction further explains the motivation of the paper and lists down the contributions and findings of the paper.

\item {\bf Limitations}
    \item[] Question: Does the paper discuss the limitations of the work performed by the authors?
    \item[] Answer: \answerYes{}
    \item[] Justification: We clearly discuss the limitations and open questions of the work in Section 6 of the paper.
    %In the end of the paper we list the limitations of our model to inform the reader.

\item {\bf Theory assumptions and proofs}
    \item[] Question: For each theoretical result, does the paper provide the full set of assumptions and a complete (and correct) proof?
    \item[] Answer: NA
    \item[] Justification: All formulation used in the paper are referenced and the empirical performances are reported in this work.

    \item {\bf Experimental result reproducibility}
    \item[] Question: Does the paper fully disclose all the information needed to reproduce the main experimental results of the paper to the extent that it affects the main claims and/or conclusions of the paper (regardless of whether the code and data are provided or not)?
    \item[] Answer: \answerYes{}
    \item[] Justification: We clarify all details about the experiments, required to reproduce the results. Some details are included in the main manuscript and others are included in the Appendix. Additionally, we open-sourced our model and code to the community.
    %In addition to the code and model that we plan to publish, we clarify to write detailed experimental setting that leaded us to achieve this results. This includes the models are formulation we use by referencing or writing, and the experimental setting such as value of the hyperparameters and etc, that needed for readers to reproduce our results. 

\item {\bf Open access to data and code}
    \item[] Question: Does the paper provide open access to the data and code, with sufficient instructions to faithfully reproduce the main experimental results, as described in supplemental material?
    \item[] Answer: \answerYes{}
    \item[] Justification: All the datasets used in this work are publicly available. We have open-sourced the model checkpoints and code. Additionally, we provide clear and detailed instructions, required to reproduce the results.
   % We plan to publish the code and all datasets we use in this work our publicly available and cited in the paper. We also mentioned the scale of the data we used in the paper.

\item {\bf Experimental setting/details}
    \item[] Question: Does the paper specify all the training and test details (e.g., data splits, hyperparameters, how they were chosen, type of optimizer, etc.) necessary to understand the results?
    \item[] Answer:  \answerYes{}
    \item[] Justification: The choice of hyperparameters and experiment details are clearly mentioned either in the main script or appendix. Full code is available publicly.

\item {\bf Experiment statistical significance}
    \item[] Question: Does the paper report error bars suitably and correctly defined or other appropriate information about the statistical significance of the experiments?
    \item[] Answer:  \answerNo{}
    \item[] Justification: As training of latent world models is quite expensive, we are not able to train each model over multiple random seeds. The models in this work are trained on large datasets and evaluated on multiple evaluation sets.
    %We conduct proper ablation studies on medium size models to justify our design choices for our large model training.
    %The evaluation 
    %The datasets and evaluation sets are large enough to produce stable measures. 
    %For sampling, we also need many samples to generate and evaluate the performance of the model. 
    %Due to the large number of samples the results would not deviate significantly over multiple rollouts. 

\item {\bf Experiments compute resources}
    \item[] Question: For each experiment, does the paper provide sufficient information on the computer resources (type of compute workers, memory, time of execution) needed to reproduce the experiments?
    \item[] Answer: \answerYes{}
    \item[] Justification: We provide the details of compute resources used for the experiments. 
    %the GPU hours we used for each experiments.
    
\item {\bf Code of ethics}
    \item[] Question: Does the research conducted in the paper conform, in every respect, with the NeurIPS Code of Ethics \url{https://neurips.cc/public/EthicsGuidelines}?
    \item[] Answer: \answerYes{}
    \item[] Justification: We Do not violate any code of ethics.

\item {\bf Broader impacts}
    \item[] Question: Does the paper discuss both potential positive societal impacts and negative societal impacts of the work performed?
    \item[] Answer: \answerYes{}
    \item[] Justification: At the end of Section 6, we discuss the social impact of our work. We discuss how this work can contribute to building reliable and cost-efficient autonomous driving systems.
    %In Section 6, we discuss the social impact of our work. 
    %And inform about the fact that our model is a generative model and requires more study before going under any use.
    
\item {\bf Safeguards}
    \item[] Question: Does the paper describe safeguards that have been put in place for responsible release of data or models that have a high risk for misuse (e.g., pretrained language models, image generators, or scraped datasets)?
    \item[] Answer: NA 
    \item[] Justification: The studied topic of research is still in its very early stage of development. %Therefore, does not pose any risk for misuse.
    %In the end of the paper, we inform the readers that our model is built upon the development of generative models and can sometimes hallucinate or do mode collapse.

\item {\bf Licenses for existing assets}
    \item[] Question: Are the creators or original owners of assets (e.g., code, data, models), used in the paper, properly credited and are the license and terms of use explicitly mentioned and properly respected?
    \item[] Answer: \answerYes{}
    \item[] Justification: We clearly cite all works, including papers and code repositories, used in this paper.

\item {\bf New assets}
    \item[] Question: Are new assets introduced in the paper well documented and is the documentation provided alongside the assets?
    \item[] Answer: \answerYes{}
    \item[] Justification: We released our code and it is well documented such that readers can reproduce the results or use provided checkpoints to get the same results as we provided in the paper.

\item {\bf Crowdsourcing and research with human subjects}
    \item[] Question: For crowdsourcing experiments and research with human subjects, does the paper include the full text of instructions given to participants and screenshots, if applicable, as well as details about compensation (if any)? 
    \item[] Answer: \answerNA{}
    \item[] Justification: This work does not involve crowdsourcing nor research with human subjects.

\item {\bf Institutional review board (IRB) approvals or equivalent for research with human subjects}
    \item[] Question: Does the paper describe potential risks incurred by study participants, whether such risks were disclosed to the subjects, and whether Institutional Review Board (IRB) approvals (or an equivalent approval/review based on the requirements of your country or institution) were obtained?
    \item[] Answer: \answerNA{}
    \item[] Justification: This work does not involve crowdsourcing nor research with human subjects.

\item {\bf Declaration of LLM usage}
    \item[] Question: Does the paper describe the usage of LLMs if it is an important, original, or non-standard component of the core methods in this research? Note that if the LLM is used only for writing, editing, or formatting purposes and does not impact the core methodology, scientific rigorousness, or originality of the research, declaration is not required.
    %this research? 
    \item[] Answer: \answerNA{}
    \item[] Justification: This work does not involve any study or usage of LLM.

\end{enumerate}

\newpage

\begin{center}
    \Large \textbf{Appendix}
\end{center}
\appendix

\section{Video Rollouts}
We include qualitative examples in video form, embedded in the \url{https://lmb-freiburg.github.io/orbis.github.io/} page.
%We include qualitative examples in video form, embedded in the Orbis-webpage. Opening the file in a web browser \textit{after extracting the whole archive} should display them in a gallery. The attached directory also contains the individual video files.
We have divided the qualitative examples into sections.

\subsection{Comparison with the state-of-the-art} Here we show videos generated by our method beside those generated by the baseline approaches (Vista~\cite{gao2024vista}, GEM~\cite{hassan2024gem}, DrivingWorld~\cite{hu2024drivingworld}, Cosmos~\cite{agarwal2025cosmos}), for the same initial condition frames.
We include videos generated by our method for both continuous (FM) and discrete (MG) version.

These videos showcase the superiority of our model in dealing with content generation after turning events. Orbis (FM) can generate more realistic scenes and objects than its discrete (MG) counterpart. Moreover, in the fifth scene our model is the only one to halt at a stop sign, generating the passing of a pedestrian and a car.

Along the generated videos, we display the estimated trajectories for Orbis, Vista, and GEM. These show the unrealistic ego-motion that the SVD-based methods produce in some cases. Trajectories are estimated using the VGGT model~\cite{wang2025vggt}.

\subsection{Performance in different scenarios} Here we show our videos generated by our model on straight drives, turns, and urban scenes. Moreover, we show how our model can generate diverse videos when starting from the same initial condition frames.

\subsection{Randomly sampled videos}
Here we show randomly sampled videos, all generated by our model, for nuPlan, Waymo, and BDD100K. The first two are out-of-domain, whereas the last is in-domain w.r.t. the model's training data.

Even though our approach can generate videos from out-of-domain condition frames, its rollouts stop more often on nuPlan and Waymo samples, compared to BDD100K.

% We compare our method against baseline approaches and show that it can perform longer rollouts while generating higher-quality frames. To demonstrate the model’s effectiveness across diverse scenarios, we present qualitative examples including straight driving, turns, and urban environments. Additionally, to avoid cherry-picking, we include randomly sampled rollouts to provide a representative view of the average generation quality.

% \begin{figure}[!ht]
%     \centering
%     \includegraphics[width=\linewidth]{pgf_plots/visualization-appendix.pdf}
%     \caption{Caption}
%     \label{fig:enter-label}
% \end{figure}

\section{Dataset Details}

% \subsection{BDD100K}
% \todo{}

\subsection{OpenDV}
We filter the training videos from OpenDV by brightness and by video title. We discard all videos containing any of the following words in their original title: \texttt{night}, \texttt{scenic}, \texttt{interstate}, \texttt{nature}, \texttt{desert}, \texttt{park}, \texttt{walking}.
We then discard all videos with an average pixel value below 90 in a [0, 255] range, in order to keep consistency with the selected BDD100K subset.

From the resulting 1337 videos we then discard the first and last 60 seconds (to avoid text and other overlays) and extract a total of 19398 30-second long clips.

\subsection{Validation Sets}
Here we describe how we obtained the validation sets used in the paper. We will release the annotation files needed to reproduce the validation sets.

\subsubsection{nuPlan}
For this benchmark, we use the validation set of nuPlan~\cite{caesar2021nuplan}, at its original sampling rate of 10Hz. We select the validation samples by ensuring a distance of 8 seconds between their starting frames and a length of at least 20 seconds worth of real frames available for evaluation.

The total resulting samples are 5878. Due to the computational cost of generating videos for all approaches we evaluate on the first 800 samples.

\subsubsection{nuPlan-turns}
For this benchmark, we use the validation set of nuPlan~\cite{caesar2021nuplan}. We select the starting frames of the validation samples based on three criteria:
\begin{itemize}
    \item a distance of at least 3 seconds between consecutive samples,
    \item at least 40 seconds worth of real frames available for evaluation,
    \item an initial yaw rate of at least 0.12 rad/s, equivalent to approximately 1 standard deviation.
\end{itemize}
We evaluate on 400 of the resulting 416 samples.

\subsubsection{Waymo}
This benchmark is based on the validation set of the Waymo Open Datset~\cite{sun2020scalability}, at its original sampling rate of 10Hz.

We select the validation samples by ensuring a distance of 2 seconds between their starting frames and a length of at least 15 seconds worth of real frames available for evaluation.
We use 400 of the resulting 406 samples selected with these criteria.

\section{Model Details}

\subsection{Latent world model: Training details}
Both continuous and discrete models follow a spatial-temporal Transformer architecture. 
ST-Transformer blocks~\cite{xu2020spatial} have interleaved spatial and temporal attention layers. 
For high-resolution experiments, we replace the spatial block with a Swin Transformer~\cite{liu2021swin}, leveraging windowed attention for efficiency. 
Our transformer architecture consists of 24 ST-blocks with a hidden dimensionality of 768, split across 16 attention heads. 
We train models with a context of 5 frames sampled at 5Hz, using the AdamW \cite{loshchilov2017decoupled} optimizer with a learning rate of $5\times10^{-5}$.

\subsection{Flow matching}

We modify the DiT~\cite{peebles2023scalable} to a STDiT architecture by decomposing temporal and spatial attention. As shown in Table~\ref{STDiT_vs_DiT}, the STDiT not only achieves a better FVD but also the frame quality, measured by FID, degrades more slowly over time.

We compute the standard deviation of the training set's encoded representations and normalize each frame by dividing by this value, ensuring unit variance across inputs ~\cite{rombach2022high}. 
This normalization occurs for detail and semantic tokens independently. To improve generalization and frame generation quality, we drop context frames 50\% of the time. 
This number reduces to 10\% after 5 epochs of training. 
When context frames are present, we augment them with noise 50\% of the time, similar to prior work~\cite{valevski2024diffusion, gao2025adaworld, he2022latent}. In order to sample the next frame, we use ODE sampler and take 30 steps \cite{lipman2022flow}.

\begin{table}[!ht]
  \caption{Comparison of DiT and STDiT performance. Metrics are computed over 200 sequences, each consisting of 120 generated frames, using the BDD100K dataset.}
  \label{STDiT_vs_DiT}
  \centering
  \begin{tabular}{lccccc}
    \toprule
    Name          & FVD~$\downarrow$     & FID~$\downarrow$ frame 30     & FID~$\downarrow$ frame 60 & FID~$\downarrow$ frame 90  & FID~$\downarrow$ frame 120 \\
    \midrule
    DiT  & 287.03     & 81.46     & 91.06   & 98.45  & 101.91   \\
    % DiT (532M)         & 285.65    & 79.05     & 84.87   & 93.02  & 94.59   \\
    STDiT  & 273.69     & 77.98     & 85.53    & 89.99  & 89.80 \\
    \bottomrule
  \end{tabular}
\end{table}

\subsection{Masked generative modeling}\label{app:sec:model_mgm}

Here, we explain the extra regularizer which is added to improve the training process of the discrete model. 

Since, the model takes discrete token indices as input from the tokenizer, it discards any pairwise similarity structure of the latent tokens.
To reintroduce this structure, we utilize the similarity matrix  $\mathbf{S} \in \mathbb{R}^{K\times K}$ over the $K$ codebook vectors and let $s_i = \mathbf{S}_i$ be the $i^{th}$ row corresponding to the ground-truth token index $i$. Formally, letting $u_i\in\mathbb R^K$ be the model's output logits for target token with index $i$, we define
\begin{equation}
p_i^o = \frac{e^{u_i/T}}{\sum_j e^{u_j/T}},\quad
p_i^t = \frac{e^{s_i/T'}}{\sum_j e^{s_j/T'}}
\end{equation}

where \(T\) and \(T'\) are temperature hyperparameters and $p_i^t$ is treated as soft-target for the model output. The objective is to minimize the KL-divergence between $p_i^o$ and $p_i^t$ as 
\begin{equation}
\mathcal{L}_{\mathrm{KD}}
= T\,T' \sum_{i=1}^{H'W'} 
      D_{\mathrm{KL}}\bigl(p^t_i \,\|\, p^o_i\bigr).
\label{eq:kd_loss}
\end{equation}

This is similar to knowledge distillation objective~\cite{hinton2015distilling}, which aims to enrich relational information by using soft-targets instead of hard one-hot labels and are known to improve data efficiency and generalization. The overall model training objective is $\mathcal{L} = \mathcal{L}_{CE} + \lambda \mathcal{L}_{KL}.$

We use $T=2$, $T'=0.2$ and $\lambda=0.5$ for our experiments.

\paragraph{Refinement module.}
The discrete masked generative model struggles to maintain temporal coherence across the full spatial extent of each frame. While it captures important temporal connections to keep the motion of objects consistent across frames and often predicts token with correct semantic property, it predicts tokens with inconsistent appearance. This is likely due to the limitations of heuristic-based unmasking scheme. % in masked generative modeling.
These inconsistencies result in flickering artifacts, which degrades the quality of the video. These artifacts negatively impacts FVD performance, especially for long-horizon prediction, where the corrupted predictions are reused as context. 
To remedy these artifacts and compare FVD fairly with continuous baselines, we introduce a small video refinement model comprising of 30 M parameters. This refinement module is only a post-processing unit and does not affect the world model learning. It follows a U-Net architecture with 12 3D-convolutional layers and operates in the latent space.
It takes four predicted frame latents from the world model as input and outputs the refined continuous latents. 

It is trained directly on the tokenizer encoder output, where it predicts clean continuous latents from the corrupted quantized tokens from the tokenizer. To simulate noise, 20\% of the quantized tokens are replaced with randomly picked top-1000 tokens based on the similarity matrix.
Ground-truth continuous latents from the hybrid image tokenizer serve as training targets.
The model is trained with a flow-matching objective to denoise corrupted latents.
At inference, refinement is applied in a sliding-window manner over 4 frame latents, sliding one frame at a time. Only the last predicted frame latent is retained and updated. We use ODE sampler and take only 1 step. The refined latents are decoded by the tokenizer to produce the final image frame.

%\section{Training details}
\begin{table}
%\small
  \centering
  \caption{Overview of the objectives used in three training phases of the hybrid image tokenizer. $\lambda_{D}=2.0, \lambda_{EQ}=0.25, \lambda_{G}=0.1$.}. 
  \label{tab:tok_phases}
  \begin{tabular}{lcccc}
      \toprule
       Total & Trainable & Train Mode & Objectives & $\#$Epochs \\
      \midrule
      Phase-1   &  Full model  & discrete-only &  $L_{rec} + L_{VQ} + L_{per} + \lambda_{D} L_{DINO}$  &  12 \\
      Phase-2   &  Full model &  discrete + cont.  & Phase-1 +  $\lambda_{EQ} L_{EQ}$  & 3 \\
      Phase-3   & Decoder-only  & discrete + cont. &   $L_{rec} + L_{per} + \lambda_{G} L_{GAN}$  & 5 \\
      \bottomrule
    \end{tabular}
  \vspace*{0.5cm}
\end{table}

\subsection{Tokenizer training details}
We initialize the VIT-Base encoder with pretrained MAE weights. Both encoder branches combined consists of 171.6 M parameters. The CNN-based decoder architecture is based on VQGAN~\cite{esser2021taming} tokenizer and consists of 44.8 M parameters. We use 16-dim latents, each for semantic and detail codebooks. 
For the final model, we train the quantized version of the image tokenizer with codebook size of 16384 for each codebook. The model training has three phases. First phase is similar to  VQGAN training, but without the adversarial loss~\cite{agarwal2025cosmos}. In the second phase, we fine-tune with scale-equivariance regularization~\cite{kouzelis2025eq}. We only fine-tune the decoder in the third phase with the adversarial loss. Three phases in total comprise of 20 epochs of training. Phase-2 and Phase-3 are trained with a mix of discrete and continuous latents (includes VQ for discrete) to enable corresponding types of world modeling, as shown in Fig. 3 of the main manuscript. In the mixed fine-tuning phases, $50\%$ mini-batches are trained with discrete latents and $50\%$ with continuous latents. Hyperparameters and objective details of three phases are included in Table~\ref{tab:tok_phases}. $L_{rec}$ refers to $L_1$ reconstruction loss, $L_{per}$ refers to perceptual loss, $L_{EQ}$ refers to scale-equivariance regularization loss, $L_{GAN}$ refers to the adversarial loss and $L_{VQ}$ refers to the vector-quantization objectives including codebook and commitment losses.  

The model is trained with a mix of 7 datasets comprising of 2.49 M images. OpenDV dataset accounts for around 90\% of the dataset. The split across all datasets included for tokenizer training is shown in Table.~\ref{tab:tokenizer-dataset}. We select only daylight images for the dataset.

%\section{Trajectory Evaluation Metrics}
\begin{table}[h!]
    \caption{Tokenizer dataset overview.}
    \label{tab:tokenizer-dataset}
    \centering
    \begin{tabular}{lc}
      \toprule
      Name         & Frames   \\
      \midrule
      OpenDV~\cite{yang2024genad}       & 2.26 M   \\
      BDD100K~\cite{yu2020bdd100k}          & 158.6 K  \\
      Honda HAD~\cite{kim2019CVPR}    & 5.1 K    \\ 
      ONCE~\cite{mao2021one}        & 14 K     \\
      Honda HDD~\cite{ramanishka2018CVPR}    & 5 K      \\
      NuScenes~\cite{Caesar_2020_CVPR}     & 3 K      \\
      NuPlan~\cite{caesar2021nuplan}      & 47.4 K   \\
      \midrule
      Total        & 2.49 M   \\
      \bottomrule
    \end{tabular}
\end{table}

%\paragraph{Evaluation Metrics}

\section{Evaluation Metrics}

\subsection{Trajectory evaluation metrics}\label{app:sec:traj_eval}

To evaluate the distributional fidelity of generated trajectories, we use two primary metrics: \textit{pointwise error} and \textit{curve similarity}. These metrics serve as distance measures to evaluate \textit{distributional fidelity and coverage} using precision--recall~\cite{kynkaanniemi2019pr} in the driving trajectory space relevant to world model evaluation. Specifically, we represent a driving trajectory as a sequence of extrinsic transformation matrices $\boldsymbol{\mathcal{T}} = (\mathbf{T}_1, \dots, \mathbf{T}_T)$, where each $\mathbf{T}_t$ comprises a rotation (orientation) $\mathbf{R}_t \in SO(3)$ and a translation (position) $\mathbf{p}_t \in \mathbb{R}^3$, arranged as $\mathbf{T}_t = [\mathbf{R}_t, \mathbf{p}_t; \mathbf{0}, 1]$. For computing the ADE and Fréchet distances, we consider only the planar positions $\mathbf{p}_t \in \mathbb{R}^2$. Other parameters within $\mathbf{T}_t$, such as the rotation $\mathbf{R}_t$, can additionally be utilized to assess realism aspects like turning behavior and orientation evolution over time.

\vspace{0.5em}
\noindent\textbf{Average Displacement Error (ADE).}
Given a predicted trajectory $\hat{\tau} = (\hat{\mathbf{p}}_1, \dots, \hat{\mathbf{p}}_T))$ and a ground-truth trajectory $\tau = (\mathbf{p}_1, \dots, \mathbf{p}_T)$, with positions $\hat{\mathbf{p}}_t, \mathbf{p}_t \in \mathbb{R}^2$, the ADE is defined as the average Euclidean distance between corresponding points:
\[
\text{ADE}(\hat{\tau}, \tau) = \frac{1}{T} \sum_{t=1}^T \|\hat{\mathbf{p}}_t - \mathbf{p}_t\|_2.
\]
This metric quantifies pointwise deviations and is sensitive to minor spatial misalignments.

\vspace{0.5em}
\noindent\textbf{Discrete Fréchet Distance.}
The discrete Fréchet distance assesses the alignment cost between two trajectories while preserving their temporal ordering:
\[
\text{FD}(\hat{\tau}, \tau) = \min_{\alpha, \beta} \max_{i = 1, \dots, m} \|\hat{\mathbf{p}}_{\alpha(i)} - \mathbf{p}_{\beta(i)}\|_2,
\]
where $\alpha,\beta$ are non-decreasing mappings from trajectory indices to points. This metric emphasizes structural similarity and penalizes shape mismatches more robustly than ADE.

\vspace{0.5em}
\noindent\textbf{Precision and Recall in Trajectory Embedding Space.}
To evaluate the distributional alignment between real and generated trajectories, we utilize the precision--recall framework~\cite{kynkaanniemi2019pr}. We first map videos into the trajectory space using an inverse dynamics model (VGGT~\cite{wang2025vggt}). Let $\mathcal{R} = \{\mathbf{r}_i\}_{i=1}^N$ and $\mathcal{G} = \{\mathbf{g}_j\}_{j=1}^M$ denote the planar positions trajectories of real and generated trajectories, respectively. For each real trajectory $\mathbf{r}_i$, define the threshold $\delta_i^{\mathcal{R}}$ as the distance to its $k$-th nearest neighbor in the real trajectory space $\mathcal{R}\setminus\{\mathbf{r}_i\}$. Conversely, for each generated trajectory $\mathbf{g}_i$, define $\delta_j^{\mathcal{G}}$ as the distance to its $k$-th nearest neighbor in the generated trajectory space $\mathcal{G}\setminus\{\mathbf{g}_j\}$. Precision and recall for a distance metric $d(\cdot,\cdot)$ (e.g., Fréchet) are then defined as:
\begin{align}
\text{Precision} &= \frac{1}{M}\sum_{j=1}^M \mathbf{1}\left[\exists\,\mathbf{r}_i \in \mathcal{R} \text{ s.t. } d(\mathbf{g}_j, \mathbf{r}_i) < \delta_i^{\mathcal{R}}\right],\\
\text{Recall} &= \frac{1}{N}\sum_{i=1}^N \mathbf{1}\left[\exists\,\mathbf{g}_j \in \mathcal{G} \text{ s.t. } d(\mathbf{r}_i, \mathbf{g}_j) < \delta_j^{\mathcal{G}}\right],
\end{align}
This adaptive, density-aware thresholding enables reliable evaluation of both fidelity (precision) and coverage (recall), offering a realistic reflection of how well the generated trajectories capture the diversity and structure of real-world driving behavior.

\subsection{VQA evaluation metrics.}
\label{app:sec:vqa}

\paragraph{PSNR and SSIM.} As shown in the Table~\ref{tab:psnr_ssim}, we computed average PSNR and SSIM metrics over two shifted video windows of 10 frames (i.e. 2 seconds at 5fps) over 400 generated videos on nuPlan-turns evaluation benchmark. We resize generated videos of all methods to same resolution for a fair comparison. We observe that Orbis model performs marginally better than other methods over the first window. However, all methods converge to similarly low numbers in the second window. 

\begin{table}[h]
  \centering
  \caption{VQA metrics: PSNR and SSIM on windowed video of 10 frames over 400 generated videos on nuPlan-turns.}
  \label{tab:psnr_ssim}
  \begin{tabular}{lcccc}
    \toprule
    & \multicolumn{2}{c}{PSNR} & \multicolumn{2}{c}{SSIM}\\
    Model & frames 0-9 & frames 10-19 & frames 0-9 & frames 10-19 \\
    \midrule
    Cosmos & 17.29 & 13.38 & 0.47 & 0.38 \\
     GEM     & 14.85 & 13.73  & 0.42 & 0.41 \\
    Vista   & 15.04 & 13.70 & 0.44 & 0.42  \\
    DW     & 17.67 & 14.70 & 0.44 & 0.38 \\
    Orbis   & 18.72 & 14.75  & 0.52 & 0.42 \\
    \bottomrule
  \end{tabular}
\end{table}

%     \begin{table}[]
%     \centering
%     \begin{tabular}{|l|c|c|c|}
%         \midrule
%         Method  & frames 0-10 & frames 10-19 \\
%         \midrule
%         Cosmos & 17.29 & 13.38  \\
%         GEM     & 14.85 & 13.73 \\
%         Vista   & 15.04 & 13.70  \\
%         DW     & 17.67 & 14.70  \\
%         Orbis   & 18.72 & 14.75  \\
%         \midrule
%     \end{tabular}
%     \caption{PSNR metrics computed over two shift windows (2 seconds) from predicted frame 0 to 9 and frame 10 to 19.}
%     \label{tab:my_label}
% \end{table}

%     \begin{table}[]
%     \centering
%     \begin{tabular}{|l|c|c|c|}
%         \midrule
%         Method  & frames 0-10 & frames 10-19 \\
%         \midrule
%         Cosmos & 0.47 & 0.38  \\
%         GEM     & 0.42 & 0.41 \\
%         Vista   & 0.44 & 0.42  \\
%         DW     & 0.44 & 0.38  \\
%         Orbis   & 0.52 & 0.42  \\
%         \midrule
%     \end{tabular}
%     \caption{SSIM metrics computed over two shift windows (2 seconds) from predicted frame 0 to 9 and frame 10 to 19.}
%     \label{tab:my_label}
% \end{table}

\paragraph{DOVER.} 
We evaluated the models on the DOVER~\cite{wu2023exploring} score, which is also used in the data curation/filtering pipeline of Cosmos. In the Table~\ref{tab:dover}, we report a comparison of the DOVER scores on the nuPlan Turns dataset, computed over 17s long videos. We include both the results for the full Cosmos pipeline (including the extra 7B-parameters diffusion refiner, "Cosmos+ref"), and for the pure Cosmos world model with a non-generative decoder.

\begin{table}[h]
  \centering
  \caption{Blind VQA DOVER metric on 400 generated videos of 17s on nuPlan-turns.}
  \label{tab:dover}
  \begin{tabular}{lc}
 \toprule
        Method & DOVER$\uparrow$ \\
        \midrule
        Cosmos & 19.94 \\
        Cosmos+refinement & 28.99 \\
        GEM    & 19.76 \\ 
        Vista  & 21.14 \\
        DW     & 21.92 \\
        Orbis  & 21.34 \\
    \bottomrule
  \end{tabular}
\end{table}

% \begin{table}[h]
%     \centering
%     \begin{tabular}{l|c}
%      \caption{DOVER scores}
%       \label{tab:dover}
%     \toprule
%         Method & DOVER$\uparrow$ \\
%         \midrule
%         Cosmos & 19.94 \\
%         Cosmos+refinement & 28.99 \\
%         GEM    & 19.76 \\ 
%         Vista  & 21.14 \\
%         DW     & 21.92 \\
%         Orbis  & 21.34 \\
%     \bottomrule
%     \end{tabular}
   
% \end{table}

\subsection{FVD evaluation.}
We compute FVD is three formats to evaluate both short and long-horizon predictions. We compute short-horizon prediction over the first 6 seconds of predicted video. Results of short-horizon are shown in Table 2 in the main manuscript and Table~\ref{tab:fvd_orig_fps} in the Appendix. 
Long-horizon FVD is evaluated in two ways: cumulative and chunked. In cumulative FVD evaluation, FVD is computed on increasing video lengths starting from 4 seconds, up to 16 seconds. Results for cumulative-FVD on nuPlan-turns dataset are shown in Fig.~\ref{fig:fvd_cumulative_turns} in the Appendix. Chunked-FVD is computed on consecutive 4 seconds windows taken at different starting timestamps, shown in Fig. 4 in the main manuscript and Fig.~\ref{fig:fvd_chunked_std} in the Appendix on nuPlan-turns and nuPlan evaluation sets respectively. 

\section{More results}\label{app:sec:more_results}

\begin{figure}[!ht]
    \centering
    \begin{subfigure}[b]{0.48\textwidth}
        \includegraphics[width=\textwidth]{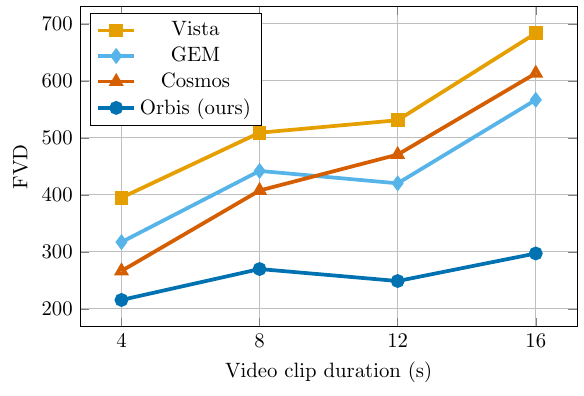}
        \caption{Video quality (FVD) over accumulated 4s time windows on nuPlan-turns. The x axis shows the video clip duration in seconds.}
        \label{fig:fvd_cumulative_turns} 
    \end{subfigure}
    \hfill
    \begin{subfigure}[b]{0.48\textwidth}
        \includegraphics[width=\textwidth]{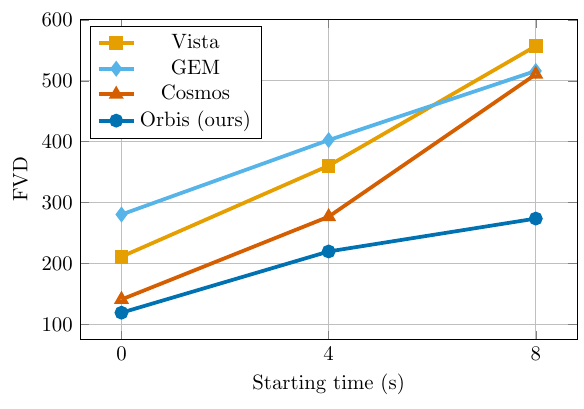}
        \caption{Video quality (FVD) over consecutive 4s time windows on standard nuPlan. The x axis shows the starting time of the evaluated time window.}
        \label{fig:fvd_chunked_std}
    \end{subfigure}
    \caption{(a) Cumulative FVD on nuPlan turns on 400 samples and (b) Chunked FVD on nuPlan standard evaluation set on 800 samples.}
    %\label{fig:mainfigure}
\end{figure}

\begin{table}[!ht]
\centering
    \caption{FVD over 6 seconds at original frame rate of different baseline methods. Lower FVD is better. *DrivingWorld (DW) is trained on the test dataset nuPlan and uses ego-motion control as an extra input.}
    \label{tab:fvd_orig_fps}
    \begin{tabular}{lccc|c}
      \toprule
      Model & fps & nuPlan & Waymo & nuPlan \\
       & &  &  &  turns\\
      \midrule
      Cosmos~\cite{agarwal2025cosmos} & 10  & 210.56 & 249.08 & 244.80 \\ 
      Vista~\cite{gao2024vista} & 10  & 289.95 & 351.42 & 353.27 \\ 
      GEM~\cite{hassan2024gem} & 10  & 348.36 & 218.61 & 318.73 \\ 
      DW*~\cite{hu2024drivingworld}  & 5 & 298.97 & N/A & 334.89\\  
      \midrule
      Orbis (ours)  & 5 & \textbf{132.25} & \textbf{180.54} & \textbf{231.88}\\ %  199.41
      \bottomrule
    \end{tabular}
\end{table}

\paragraph{FVD at original frame rate.}
Originally, the previously published models were trained and evaluated with different frame rates.
The main manuscript evaluated all models at 5hz for a fair comparison, skipping alternative frames if the prediction frame rate is 10hz. 
Here, we also include FVD scores at original prediction frame rates over 6 seconds rollouts, shown in Table~\ref{tab:fvd_orig_fps}. 
The models evaluated at 10hz achieve lower FVD scores than their 5hz counterparts. 
Despite FVD’s sensitivity to frame rate, our model at 5hz still outperforms prior approaches evaluated at higher frame rates.

\paragraph{Cumulative FVD on nuPlan-turns.}
Figure~\ref{fig:fvd_cumulative_turns} shows results for cumulative-FVD scores on the nuPlan-turns evaluation set. Our proposed model consistently outperforms other baselines, showing a strong scalable behavior as the prediction window extends.

\paragraph{Chunked FVD on nuPlan.}
We also evaluate chunked FVD on nuPlan evaluation set using 800 samples, shown in Fig.~\ref{fig:fvd_chunked_std}. Our model consistently outperforms all baseline across all video windows. Cosmos performs relatively well on early prediction windows but degrades very quickly over later windows. In contrast, GEM performs worse in early windows, but extends well for later windows. We observe GEM suffers in the early prediction windows likely due to its single frame context, which causes it to deviate from the ground truth trajectory earlier than other baselines. However, GEM generates better content in later windows, outperforming other baselines over extended predictions.

\begin{figure}[h!]
    \centering
    \includegraphics[width=0.5\linewidth]{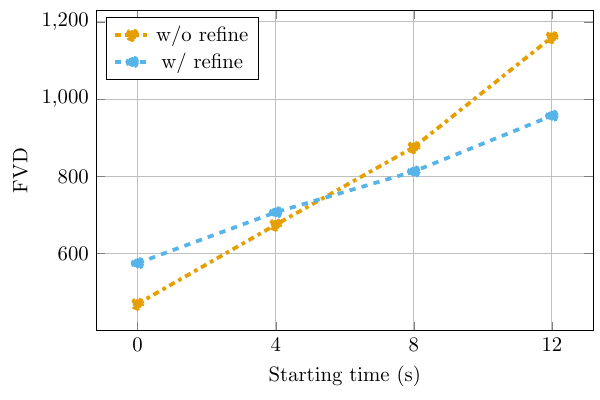}
    \caption{Effect of refinement module in masked generative modeling (orbis-MG model). Video quality (FVD) over consecutive 4s time windows on nuPlan-turns. The x axis shows the starting time of the evaluated time window.}
    \label{fig:FVD_chunked_refine}
\end{figure}

\paragraph{Effect of refinement module.}
The refinement module is design to reduce flickering artifacts caused by imprecise decoding of frames in masked generated modeling. We find that refinement module is effective for long-horizon predictions, where the context is usually corrupted. However, the module has a detrimental effect on short-horizon performance. Fig.~\ref{fig:FVD_chunked_refine} shows FVD on nuPlan-turns in a windowed (chunked) evaluation, with and without the usage of refinement module. We observe that the refinement module shows improvement for long-horizon prediction, especially longer than 6 seconds.

\paragraph{Inference speed and memory requirements.} 
Table~\ref{tab:inf_speed} compares the average times (fps) needed to generate a frame, and the required GPU memory. Orbis has the best throughput compared to all other methods. This advantage can be attributed to the smaller size of the Orbis model. By parallelizing Orbis' computation in batches we can achieve an higher
throughput. GEM and Vista are based on the same architecture but use different sampling protocols trading off FPS and VRAM.

We also compare the inference speed of our discrete Orbis-MG model and the
continuous Orbis-FM model. We also report the GPU (VRAM) memory requirements for both methods during inference. Orbis-MG shows better inference speed  compared to the Orbis-FM model. However, since Orbis-FM achieves significantly better video generation performance, it remains the default choice despite the speed advantage of the discrete model.

\begin{table}[h]
\centering
    \caption{Comparison of inference speed and VRAM memory requirements of different models.}
    \label{tab:inf_speed}
    \begin{tabular}{l|c|c}
        \toprule
        Method  & FPS$\uparrow$ & VRAM (GB) $\downarrow$ \\
        \midrule
        Cosmos  & 0.18 & 29  \\
        Vista   & 0.58 & 86  \\
        GEM     & 0.44 & 45  \\
        DW      & 0.25 & 10  \\
        Orbis-FM (ours)   & 0.70 & 24  \\
        Orbis-MG (ours)   & 0.85 & 21  \\
        \bottomrule
    \end{tabular}
\end{table}

\end{document}